\documentclass{article}

\usepackage[preprint]{Convex_NN}

\usepackage[utf8]{inputenc} 
\usepackage[T1]{fontenc}    
\usepackage{hyperref}       
\usepackage{url}            
\usepackage{booktabs}       
\usepackage{amsfonts}       
\usepackage{nicefrac}       
\usepackage{microtype}      
\usepackage{xcolor}         
\usepackage{algorithm,algorithmic}
\usepackage{adjustbox}
\usepackage{caption,subcaption}
\usepackage{enumitem}
\usepackage{authblk}

\usepackage{amsmath,amsfonts,bm}
\usepackage{mathtools}
\usepackage{amssymb}
\usepackage{amsthm}

\def\eqref#1{equation~\ref{#1}}

\def\1{\bm{1}}

\def\eps{{\epsilon}}

\DeclareMathAlphabet{\mathsfit}{\encodingdefault}{\sfdefault}{m}{sl}
\SetMathAlphabet{\mathsfit}{bold}{\encodingdefault}{\sfdefault}{bx}{n}

\def\gD{{\mathcal{D}}}

\def\gF{{\mathcal{F}}}

\def\gN{{\mathcal{N}}}
\def\gO{{\mathcal{O}}}

\def\gS{{\mathcal{S}}}

\def\gU{{\mathcal{U}}}

\def\gX{{\mathcal{X}}}

\def\sP{{\mathbb{P}}}

\def\sR{{\mathbb{R}}}

\newcommand{\R}{\mathbb{R}}

\DeclareMathOperator*{\argmax}{arg\,max}

\newcommand{\diag}{\text{diag}}
\newcommand{\sgn}{\text{sgn}}

\DeclarePairedDelimiter{\norm}{\lVert}{\rVert}

\DeclarePairedDelimiter\biggnorm{\bigg\lVert}{\bigg\rVert}
\DeclarePairedDelimiter{\enorm}{\lVert}{\rVert_2}
\DeclareMathOperator{\ST}{s. t. }
\DeclarePairedDelimiter{\rbr}{(}{)}

\newcommand{\jm}{_{j = 1}^m}
\newcommand{\jms}{_{j = 1}^{m^\star}}
\newcommand{\mhs}{\widehat{m}^\star}
\newcommand{\jmh}{_{j = 1}^{\widehat{m}}}
\newcommand{\jmhs}{_{j = 1}^{\mhs}}

\newcommand{\iP}{_{i = 1}^P}

\newcommand{\itildeP}{_{i = 1}^{\widetilde P}}

\newcommand{\iPhat}{_{i = 1}^{\widehat{P}}}
\newcommand{\iPs}{_{i = 1}^{P_s}}
\newcommand{\iPss}{_{i = 1}^{P_s + 1}}

\newcommand{\allDelta}{_{\Delta: X+\Delta \in \gX}}
\newcommand{\allDeltaU}{_{\Delta: X+\Delta \in \gU}}

\newcommand{\alliinP}{\forall i \in [P]}
\newcommand{\alliinPs}{\forall i \in [P_s]}
\newcommand{\alliinPss}{\forall i \in [P_s + 1]}
\newcommand{\alliinPhat}{\forall i \in [\widehat{P}]}

\newcommand{\one}{\mathbf{1}}

\newcommand{\dequal}{\overset{\Delta}{=}}

\newtheorem{theorem}{Theorem}
\newtheorem*{theorem*}{Theorem}
\newtheorem{corollary}{Corollary}[theorem]
\newtheorem*{corollary*}{Corollary}
\newtheorem{lemma}[theorem]{Lemma}

\title{Practical Convex Formulation of Robust One-hidden-layer Neural Network Training}

\newcommand{\ybemail}{\normalfont{\href{mailto:Yatong Bai <yatong_bai@berkeley.edu>?Subject=Your NeurIPS 2021 paper}{\texttt{yatong\_bai@berkeley.edu}}}}
\newcommand{\tgemail}{\normalfont{\href{mailto:Tanmay Gautam <tgautam23@berkeley.edu>?Subject=Your NeurIPS 2021 paper}{\texttt{tgautam23@berkeley.edu}}}}
\newcommand{\ygemail}{\normalfont{\href{mailto:Yu Gai <yu_gai@berkeley.edu>?Subject=Your NeurIPS 2021 paper}{\texttt{yu\_gai@berkeley.edu}}}}
\newcommand{\ssemail}{\normalfont{\href{mailto:Somayeh Sojoudi <sojoudi@berkeley.edu>?Subject=Your NeurIPS 2021 paper}{\texttt{sojoudi@berkeley.edu}}}}

\begin{document}

\author[1]{\textbf{Yatong Bai}}
\author[2]{\textbf{Tanmay Gautam}}
\author[2]{\textbf{Yu Gai}}
\author[1,2]{\textbf{Somayeh Sojoudi}}

\affil[1]{Department of Mechanical Engineering, University of California, Berkeley}
\affil[2]{Department of Electrical Engineering and Computer Science, University of California, Berkeley}
\affil[ ]{ }
\affil[ ]{\ybemail \quad \tgemail}
\affil[ ]{\;\; \ygemail \qquad\; \ssemail}

\maketitle

\begin{abstract}
    Recent work has shown that the training of a one-hidden-layer, scalar-output fully-connected ReLU neural network can be reformulated as a finite-dimensional convex program.
    Unfortunately, the scale of such a convex program grows exponentially in data size. In this work, we prove that a stochastic procedure with a linear complexity well approximates the exact formulation.
    Moreover, we derive a convex optimization approach to efficiently solve the ``adversarial training'' problem, which trains neural networks that are robust to adversarial input perturbations. Our method can be applied to binary classification and regression, and provides an alternative to the current adversarial training methods, such as Fast Gradient Sign Method (FGSM) and Projected Gradient Descent (PGD).
    We demonstrate in experiments that the proposed method achieves a noticeably better adversarial robustness and performance than the existing methods.
\end{abstract}

\section{Introduction} \label{sec:intro}
\label{introduction}
Over the past decade, deep learning has become one of the most prominent subfields of machine learning. Neural networks are used in various applications, ranging from natural language processing to computer vision and reinforcement learning \citep{madry2018towards,Goodfellow-et-al-2016,NIPS2012_c399862d,DBLP:journals/corr/MnihKSGAWR13}. They are also studied for safety-critical systems, such as autonomous driving \citep{DBLP:journals/corr/BojarskiTDFFGJM16}. As neural networks form the backbone of modern-day technology, it is critical to guarantee their safety while accelerating their training. One-hidden-layer neural networks are the simplest form possessing the vast representation power of neural networks \citep{DBLP:journals/nn/Hornik91} and their theoretical analysis helps with understanding more complex networks \citep{du2018gradient, JMLR:v20:18-674}.

Adversarial robustness has emerged as a powerful framework for evaluating the safety of machine learning models \citep{madry2018towards}. In the field of computer vision, for instance, it has been shown that slight manipulations in the input images can elicit misclassifications in neural networks with high confidence \citep{DBLP:journals/corr/SzegedyZSBEGF13, DBLP:conf/cvpr/Moosavi-Dezfooli16,DBLP:journals/corr/GoodfellowSS14}. Thus, adversarial robustness is crucial to safety-critical technologies such as autonomous driving, where highly susceptible models could result in grave consequences \citep{DBLP:conf/iclr/KurakinGB17}.

\subsection{Related work and overview}
The training of neural networks usually relies on Stochastic Gradient Descent (SGD), which only guarantees convergence to a local minimum for non-convex programs, including widely-used neural network training formulations. While it has been shown that gradient descent can converge to a global optimizer for one-hidden-layer ReLU networks when they are wide enough \citep{DBLP:journals/corr/abs-2006-05900,du2018gradient} or when the inputs follow a Gaussian distribution \citep{DBLP:conf/icml/BrutzkusG17}, spurious local minima can still exist in general applications.

Convex programs have the nice property that all local minima are global. To overcome the issue of arriving at spurious local minima when training neural networks, existing works have considered convexifying the neural network training problem \citep{NIPS2005_0fc170ec,JMLR:v18:14-546}. More recently, \citep{pmlr-v119-pilanci20a} proposed a convex optimization problem with the same global minimum as the non-convex cost function for a one-hidden-layer fully-connected ReLU neural network. While the explicit focus is on the case of squared loss, their analysis extends to arbitrary convex loss functions.

Unfortunately, the size of the convex program proposed in \citep{pmlr-v119-pilanci20a} grows exponentially with respect to the rank of the training data matrix, leading to an exponential overall complexity. To address this issue, we analyze a practical stochastic approximation procedure that forms convex training programs whose scales grow linearly with respect to the training data size. We provide a bound on the level of suboptimality of this approximation and empirically show that the convex approximation procedure returns a lower training cost than directly minimizing the non-convex cost function with SGD back-propagation.

On the adversarial robustness side, while there have been studies on robustness certification \citep{DBLP:conf/cdc/AndersonMLS20,DBLP:journals/corr/abs-2010-08603}, researchers have also been working extensively on training classifiers whose predictions are robust to input perturbations \citep{DBLP:conf/iclr/KurakinGB17,DBLP:journals/corr/GoodfellowSS14,DBLP:journals/corr/HuangXSS15}. ``Adversarial training'' is one of the most effective methods to train robust classifiers, compared with other methods such as obfuscated gradients \citep{pmlr-v80-athalye18a}. More recently, \citep{pmlr-v97-cohen19c} analyzed the feasibility of achieving robustness via ``random smoothing''. However, this method is more suitable for defending $\ell_2$ attacks rather than the more common $\ell_\infty$ attacks \citep{JMLR:v21:20-209}.

Optimizing the adversarial training cost function requires solving a highly non-convex minimax problem, which is difficult to solve efficiently. In this work, we tackle the issue by building upon aforementioned works to develop convex robust optimization problems for adversarial training, specifically focusing on the cases of hinge loss (for binary classification) and squared loss (for regression). We also offer experiment results to demonstrate the efficacy of the proposed adversarial training formulation and its advantages over the traditional methods.

\section{Background} \label{sec:PF}

\subsection{Notations}
Throughout this work, we focus on fully-connected neural networks with one ReLU-activated hidden layer and a scalar output, defined as 
\vspace{-1.5mm}
\begin{equation*}
    \widehat{y} = \sum\jm (X u_j)_+ \alpha_j,
\end{equation*} \\[-1.5mm]
where $X \in \sR^{n \times d}$ is the input data matrix with $n$ data points in $\sR^d$ and $\widehat{y} \in \R^n$ is the output vector of the neural network. We denote the target output used for training as $y\in\R^n$. $u_1, \ldots, u_m \in \sR^d$ are the weight vectors of the $m$ neurons in the hidden layer while $\alpha_1, \ldots, \alpha_m \in \sR$ are the weights of the output layer. The symbol $(\cdot)_+ = \max \{0, \cdot\}$ indicates the ReLU activation function.

Furthermore, let $\norm{\cdot}_p$ denote the $\ell_p$-norm within $\R^n$ and $\odot$ denote the Hadamard product. For $P \in \mathbb{N}_+$, we define $[P]$ as the set $\{a \in \mathbb{N}_+ | a \leq P\}$, where $\mathbb{N}_+$ is the set of positive integer numbers. For $q \in \R^n$, sgn$(q)\in\R^n$ denotes the sign of each entry of $q$. $[q \geq 0]$ denotes a boolean vector in $\{0,1\}^n$ with ones at the locations of the nonnegative entries of $q$ and zeros at the remaining locations. The symbol $\diag(q)$ denotes a diagonal matrix $Q \in \R^{n \times n}$, where $Q_{ii} = q_i$ for all $i$, and $Q_{ij} = 0$ for all $i \neq j$. The symbol $\one$ defines a column vector with all entries being 1. For $a\in\R^n$ and $b\in\R$, the inequality $a \geq b$ means that $a_i \geq b$ for all $i\in[n]$. For a set $\gS$, the notation $\Pi_\gS(\cdot)$ denotes the projection onto the set and $|\gS|$ denotes the cardinality of the set. For a random variable $r\in\R^n$, the notation $r\sim\gN(0,I_n)$ indicates that $r$ is a standard normal random vector.

\subsection{Convex neural-network training}

We define the problem of training the above neural network with a regularized convex loss function $\ell(\widehat{y},y)$ as:
\begin{equation} \label{nonconvex_general}
    \min_{(u_j, \alpha_j)\jm} \ell \bigg( \sum\jm (X u_j)_+ \alpha_j, y \bigg) + \frac{\beta}{2} \sum\jm \big( \norm{u_j}_2^2 + \alpha_j^2 \big).
\end{equation}
where $\beta > 0$ is a regularization parameter.
Consider a set of diagonal matrices $\{ \diag([Xu\geq 0]) | u\in\R^d \}$, and let the distinct elements of this set be denoted as $D_1, \dots, D_P$. The constant $P$ corresponds to the total number of partitions of $\R^d$ by hyperplanes passing through the origin that are also perpendicular to the rows of $X$ \citep{pmlr-v119-pilanci20a}. 
Intuitively, $P$ can be regarded as the number of possible ReLU activation patterns associated with $X$.

Consider the convex optimization problem
\begin{align}\label{convex_general}
    \min_{(v_i, w_i)\iP} & \ell \bigg( \sum_{i=1}^P D_i X (v_i-w_i), y \bigg) + \beta \sum\iP \Big( \norm{v_i}_2 + \norm{w_i}_2 \Big) \\
    \ST \quad & (2D_i-I_n) X v_i\geq 0, \; (2D_i-I_n) X w_i\geq 0, \quad \alliinP \nonumber
\end{align}
and its dual formulation
\begin{align} \label{eq:dual}
    & \max_v -\ell^*(v) \qquad \ST \;\; | v^\top (Xu)_+ | \leq \beta, \;\; \forall u: \norm{u}_2 \leq 1
\end{align}
where $\ell^*(v) \dequal \max_z z^\top v - \ell(z,y)$ is the Fenchel conjugate function. Note that (\ref{eq:dual}) is a convex semi-infinite program. 
The next theorem borrowed from \citep{pmlr-v119-pilanci20a} explains the relationship between the non-convex training problem (\ref{nonconvex_general}), the convex problem (\ref{convex_general}), and the dual problem (\ref{eq:dual}) when the neural network is sufficiently wide. 

\vspace{1mm}
\begin{theorem}[\citep{pmlr-v119-pilanci20a}] \label{THM:PILANCI}
    Let $(v_i^\star,w_i^\star)\iP$ denote a solution of (\ref{convex_general}) and define $m^\star$ as $|\{i : v_i^\star \neq 0\}| + |\{i : w_i^\star \neq 0\}|$. Given a convex loss function $\ell (\cdot, y)$, the non-convex problem (\ref{nonconvex_general}) has the same optimal objective as the convex problem (\ref{convex_general}) provided that the neural network width $m$ is at least $m^\star$, where $m^\star$ is upper-bounded by $n+1$. 
    Moreover, (\ref{eq:dual}) is a strong dual to (\ref{nonconvex_general}) and also attains the same optimal objective. The optimal neural network weights $(u_j^\star, \alpha_j^\star)\jm$ can be recovered using the formulas
    \begin{equation}\label{recover_weights} 
        \begin{aligned}
            (u_{j_{1 i}}^\star, \alpha_{j_{1 i}}^\star) & = \Big( \dfrac{v_i^\star}{\sqrt{\norm{v_i^\star}_2}}, \sqrt{\norm{v_i^\star}_2} \Big) \hspace{8mm} \text{if $v_i^\star \neq 0$}; \\
            (u_{j_{2 i}}^\star, \alpha_{j_{2 i}}^\star) & = \Big( \dfrac{w_i^\star}{\sqrt{\norm{w_i^\star}_2}}, -\sqrt{\norm{w_i^\star}_2} \Big) \quad \text{if $w_i^\star \neq 0$}.
        \end{aligned}
    \end{equation}
    where the remaining $m-m^\star$ neurons are chosen to have zero weights.
\end{theorem}

\subsection{Adversarial training} \label{adv_train_section}

\citep{DBLP:journals/corr/GoodfellowSS14} proposes that a classifier is considered robust against adversarial perturbations if it assigns the same label to all inputs within an $\ell_\infty$ bound with radius $\epsilon$. The uncertainty set can then be defined as
\begin{align} \label{perturbation_set}
    \gX = \Big\{ & X + \Delta \in \sR^{n \times d} \ \Big| \ \Delta = [\delta_1, \ldots, \delta_n]^\top, \delta_k\in\R^d, \norm{\delta_k}_\infty \leq \epsilon, \forall k \in [n] \Big\}. \nonumber
\end{align}

As suggested in \citep{madry2018towards}, one common method for training robust classifiers is to minimize the maximum loss within the perturbation set by solving the following minimax problem:
\begin{equation}\label{robust_general}
\begin{aligned}
        \min_{(u_j, \alpha_j)\jm} 
        \begin{pmatrix}
            \displaystyle \max\allDelta \ell \bigg( \sum_{j=1}^m \big( (X+\Delta) u_j \big)_+ \alpha_j, y \bigg) + \frac{\beta}{2} \sum_{j=1}^m \big( \norm{u_j}_2^2 + \alpha_j^2 \big) \hfill
        \end{pmatrix}
    \end{aligned}
\end{equation}

This process of ``training with adversarial data'' is often referred to as ``adversarial training'', as opposed to ``standard training'' that trains on clean data. In the prior literature, Fast Gradient Sign Method (FGSM) and Projected Gradient Descent (PGD) are commonly used to numerically solve the inner maximization of (\ref{robust_general}) and generate adversarial examples in practice \citep{madry2018towards}. More specifically, FGSM generates adversarial examples $\tilde{x}$ using
\vspace{-.5mm}
\begin{equation}\label{FGSM}
    \tilde{x} = x + \epsilon \cdot \sgn \Big( \nabla_x \ell \big( \sum_{j=1}^m (x^\top u_j)_+ \alpha_j, y \big) \Big).
\end{equation}

Since FGSM is a one-shot method that assumes linearity, it may miss the worst-case adversarial input. PGD better explores the nonlinear landscape of the problem and is capable of generating ``universal'' first-order adversaries by running the iterations
\begin{equation}\label{PGD}
    \tilde{x}^{t+1} = \Pi_{\gX} \bigg( \tilde{x}^t + \gamma \cdot \sgn \Big( \nabla_x \ell \big( \sum\jm (x^\top u_j)_+ \alpha_j, y \big) \Big) \bigg), \quad \tilde{x}^0 = x
\end{equation}
for $t=0,1,\dots$, where $x^t$ is the perturbed data vector at iteration $t$, $\Pi_\gX$ denotes the projection onto the set $\gX$, and $\gamma > 0$ is the step size.

\section{Practical Convex Training}

The worst-case computational complexity of solving (\ref{convex_general}) for the case of squared loss is $\gO \big( d^3 r^3 (\frac{n}{r})^{3 r} \big)$ using standard interior-point solvers \citep{pmlr-v119-pilanci20a}. Here, $r$ is the rank of the data matrix $X$ and in many cases $r=d$. 
Such a complexity is polynomial in $n$, a significant improvement over previous algorithms, but is exponential in $r$ and is thus prohibitively high for many practical applications.
For example, $d = 3 \times 32 \times 32 = 3072$ for the CIFAR-10 and CIFAR-100 image classification datasets. Such high complexity is due to the large number of $D_i$ matrices, which is upper-bounded by $2r \big( \frac{e(n-1)}{r} \big)^r$ \citep{pmlr-v119-pilanci20a}. 
To tackle this issue, we introduce Algorithm \ref{alg:train} that approximately solves (\ref{convex_general}) by independently sampling a subset of the $D_i$ matrices. Alg \ref{alg:train} can train networks with widths much less than $m^\star$.

\begin{algorithm}
    \begin{algorithmic}[1]
        \STATE Generate $P_s$ distinct diagonal matrices via $D_i \leftarrow \diag ([X a_i \geq 0])$, where $a_i \sim \gN (0, I_d)$ i.i.d. for all $i\in[P_s]$.
        \STATE \text{Solve} \vspace{-4mm}
        \begin{align} \label{eq:prac_clean}
            p_{s1}^\star = \min_{(v_i, w_i)\iPs} & \ell \Big( \sum\iPs D_i X (v_i-w_i), y \Big) + \beta \sum\iPs \big( \norm{v_i}_2 + \norm{w_i}_2 \big) \\
            \ST \quad & (2D_i-I_n) X v_i\geq 0, \; (2D_i-I_n) X w_i\geq 0, \quad \alliinPs. \nonumber
        \end{align}; \vspace{-4mm}
        \STATE \text{Recover $u_1, \ldots, u_{m_s}$ and $\alpha_1, \ldots, \alpha_{m_s}$ from the solution} \text{$(v_{s_i}^\star,w_{s_i}^\star)_{i=1}^{P_s}$ of (\ref{eq:prac_clean}) using (\ref{recover_weights})}.
    \end{algorithmic}
    \caption{Practical training}
    \label{alg:train}
\end{algorithm} 

The following theorem provides a probabilistic bound on the level of suboptimality of the neural network trained using Alg \ref{alg:train}.

\begin{theorem} \label{thm:prac}
    Consider an additional diagonal matrix $D_{P_s + 1}$ sampled independently via the process described in Alg \ref{alg:train}, and then construct 
    \vspace{-1.5mm}
    \begin{align} \label{eq:prac_clean2}
        p_{s2}^\star = \min_{(v_i, w_i)\iPss} & \ell \Big( \sum\iPss D_i X (v_i-w_i), y \Big) + \beta \sum\iPss \big( \norm{v_i}_2 + \norm{w_i}_2 \big) \\
        \ST \quad & (2D_i-I_n) X v_i\geq 0, \; (2D_i-I_n) X w_i\geq 0, \;\; \alliinPss. \nonumber
        \vspace{-3.5mm}
    \end{align}
    If $P_s \geq \frac{n+1}{\psi \xi} - 1$, where $\psi$ and $\xi$ are preset confidence level constants between 0 and 1, then with probability no smaller than $1-\xi$, it holds that $\sP \{ p_{s2}^\star < p_{s1}^\star \} \leq \psi$.
\end{theorem}

The proof of Theorem \ref{thm:prac} is presented in section \ref{sec:pracproof}. Intuitively, Theorem \ref{thm:prac} shows that independently sampling an additional $D_{P_s + 1}$ matrix will not reduce the training cost with high probability.

Compared with the exponential relationship between $P$ and $r$, a satisfactory value of $P_s$ should be on the order of $\frac{n}{\xi\phi}$ and therefore it has a linear relationship with $n$ and is independent of $r$. Thus, when $r$ is large, solving the approximated formulation (\ref{eq:prac_clean}) is significantly (exponentially) more efficient than solving the exact formulation (\ref{convex_general}). On the other hand, Alg \ref{alg:train} is no longer deterministic due to the stochastic sampling of the $D_i$ matrices, and yields solutions that upper-bound those of (\ref{convex_general}). While Alg \ref{alg:train} is not exact, we have verified empirically (shown in Appendix \ref{sec:prac_exp}) that even when $P_s$ is significantly smaller than $P$, Alg \ref{alg:train} still reliably returns a low training cost.

\section{Convex Adversarial Training} \label{sec:general_robust}
While PGD adversaries have been considered as ``universal'', adversarial training with PGD adversaries has several limitations. Since the optimization landscapes of neural networks are generally non-concave over $\Delta$, there is no guarantee that PGD will find the true worst-case adversary within the perturbation bound. Our experiments show that back propagation gradient methods can struggle to solve (\ref{robust_general}) and can be very sensitive to initializations. Moreover, iteratively solving the bi-level optimization (\ref{robust_general}) requires an algorithm with a nested loop structure, which is computationally cumbersome. To conquer such difficulties, we leverage Theorem \ref{THM:PILANCI} to re-characterize (\ref{robust_general}) as robust, convex upper-bound problems that can be efficiently minimized globally.

We first develop a result about adversarial training involving general convex loss functions. The proof is provided in section \ref{sec:CVX_MINIMAX}.
Consider the optimization problem
\refstepcounter{equation}\label{eq:rob_gen_cvx}
\begin{align}
    & \min_{(v_i, w_i)\iPhat} 
    \begin{pmatrix}
        \displaystyle \max\allDeltaU \ell \bigg( \sum\iPhat D_i (X+\Delta) (v_i - w_i), y \bigg) + \ \beta \sum\iPhat \big( \norm{v_i}_2 + \norm{w_i}_2 \big) \hfill
    \end{pmatrix} \tag{10a} \label{eq:rob_gen_cvx_1} \\
    \ST \; & \min\allDeltaU (2 D_i - I_n) (X+\Delta) v_i \geq 0, 
    \; \min\allDeltaU (2 D_i - I_n) (X+\Delta) w_i \geq 0, \;\; \alliinPhat \tag{10b} \label{eq:rob_gen_cvx_2}
\end{align}
where $\gU$ is any convex additive perturbation set and $D_1$, $\dots$, $D_{\widehat{P}}$ denote all distinct diagonal matrices $\diag([(X+\Delta) u \geq 0])$ that can be obtained for all $u \in \R^d$ and all $\Delta: X+\Delta \in \gU$ (note that the number of such matrices is shown by $\widehat{P}$).

\vspace{.25mm}
\begin{theorem} 
\label{THM:CVX_MINIMAX}
    Let $(v_{\text{rob}_i}^\star, w_{\text{rob}_i}^\star)\iPhat$ denote a solution of (\ref{eq:rob_gen_cvx}) and define $\mhs$ as $|\{i : v_{\text{rob}_i}^\star \neq 0\}| + |\{i : w_{\text{rob}_i}^\star \neq 0\}|$. When the neural network width $m$ satisfies $m \geq \mhs$, the optimization problem (\ref{eq:rob_gen_cvx}) provides an upper-bound on the non-convex adversarial training problem (\ref{robust_general}). The robust neural network weights $(u_{\text{rob}_j}^\star, \alpha_{\text{rob}_j}^\star)\jmh$ can be recovered using (\ref{recover_weights}).
\end{theorem}

When the uncertainty set is zero, Theorem \ref{THM:CVX_MINIMAX} reduces to Theorem \ref{THM:PILANCI}. In light of Theorem \ref{THM:CVX_MINIMAX}, we use optimization (\ref{eq:rob_gen_cvx}) as a surrogate for optimization (\ref{robust_general}) to train the neural network. We will show that the new problem can be efficiently solved in important cases. By the analogy to Theorem \ref{thm:prac}, an approximation to (\ref{eq:rob_gen_cvx}) can be applied to train neural networks with width much less than $\mhs$.
Since (\ref{eq:rob_gen_cvx}) includes all $D_i$ matrices in (\ref{convex_general}), we have $\widehat{P} \geq P$. While $\widehat{P}$ is at most $2^n$ in the worst case, since $\epsilon$ is often small, we expect $\widehat{P}$ to be relatively close to $P$, where $P \leq 2r \big( \frac{e(n-1)}{r} \big)^r$ as discussed above. 

The robust constraints in (\ref{eq:rob_gen_cvx_2}) force all points within the perturbation set to be feasible. 
Intuitively, for every $j\in[\widehat{m}^\star]$, (\ref{eq:rob_gen_cvx_2}) forces the ReLU activation pattern sgn$\big( (X+\Delta) u_{\text{rob}_j}^\star \big)$ to stay the same for all $\Delta$ such that $X+\Delta\in\gU$. Moreover, if $\Delta_{\text{rob}}^\star$ denote a solution to the inner maximization in (\ref{eq:rob_gen_cvx_1}), then $X + \Delta_{\text{rob}}^\star$ corresponds to the worst-case adversarial inputs for the recovered neural network.

\vspace{1.5mm}
\begin{corollary}
\label{CORO:ROB_CONSTRAINT}
    For the perturbation set $\gX$, the constraints in (\ref{eq:rob_gen_cvx_2}) can be equivalently replaced by 
    \begin{equation}\label{robust_constraint}
    \begin{aligned}
        (2 D_i - I_n) X v_i \geq \epsilon \norm{v_i}_1, \quad
        (2 D_i - I_n) X w_i \geq \epsilon \norm{w_i}_1, \quad & \alliinPhat
    \end{aligned}
    \end{equation}
\end{corollary}
The proof of the corollary is provided in section \ref{sec:ROB_CONSTRAINT}. Each vector element in the left-hand-sides should be greater than or equal to the corresponding scalar in the right-hand-sides.

\section{Convex Hinge Loss Adversarial Training}\label{sec:hinge}

While the inner maximization of the robust problem (\ref{eq:rob_gen_cvx}) is still hard to solve in general, it is tractable for some loss functions. The simplest case is the piecewise-linear hinge loss $\ell(\widehat{y}, y) = (1-\widehat{y} \odot y)_+$, which is widely used for classification. Here, we focus on binary classification with $y \in \{-1, 1\}^n$.\footnote{
Other $\ell_p$ norm-bounded additive perturbation sets can be similarly analyzed, as shown in \ref{lpnorm}. It is also straightforward to extend the analysis in this section to any convex piecewise-affine loss functions.}

Consider the training problem for a one-hidden-layer neural network with $\ell_2$ regularized hinge loss:
\begin{equation} \label{hinge_std}
    \min_{(u_j, \alpha_j)\jm}
    \bigg( \frac{1}{n} \cdot \one^\top \Big( \one - y \odot \sum\jm (X u_j)_+ \alpha_j \Big)_+ + \frac{\beta}{2} \sum\jm \big( \norm{u_j}_2^2 + \alpha_j^2 \big) \bigg)
\end{equation}

The adversarial training problem considering the $\ell_\infty$-bounded adversarial data uncertainty $\gX$ is:
\begin{align} \label{hinge_adv}
    & \min_{(u_j, \alpha_j)\jm} 
    \begin{pmatrix}
        \displaystyle \max\allDelta \frac{1}{n} \cdot \one^\top \bigg( \one - y \odot \sum\jm \big( (X+\Delta) u_j \big)_+ \alpha_j \bigg)_+ + \frac{\beta}{2} \sum\jm \big( \norm{u_j}_2^2 + \alpha_j^2 \big) \hfill \\
    \end{pmatrix}
\end{align}

Applying Theorem \ref{THM:CVX_MINIMAX} and Corollary \ref{CORO:ROB_CONSTRAINT} leads to the following formulation as an upper bound on (\ref{hinge_adv}):
\begin{align}\label{HINGE_ADV_D_minimax}
    \min_{(v_i, w_i)\iPhat} & 
    \begin{pmatrix}
        \displaystyle \max\allDelta \frac{1}{n} \cdot \one^\top \bigg( \one - y \odot \sum\iPhat D_i (X+\Delta) (v_i - w_i) \bigg)_+ + \beta \sum\iPhat \big( \norm{v_i}_2 + \norm{w_i}_2 \big) \hfill
    \end{pmatrix} \nonumber \\
    \ST \;\; & (2 D_i - I_n) X v_i \geq \epsilon \norm{v_i}_1, \;\;
    (2 D_i - I_n) X w_i \geq \epsilon \norm{w_i}_1, \;\; \alliinPhat
\end{align}

Instead of enumerating an infinite number of points in $\gX$, we only need to enumerate all vertices of $\gX$, which is finite. This is because the solution $\Delta_\text{hinge}^\star$ to the inner maximum always occurs at a vertex of $\gX$, as will be shown in Theorem \ref{LEMMA:INNER_MAX}. 
Solving the inner maximization of (\ref{HINGE_ADV_D_minimax}) in closed form leads us to the next theorem, whose proof is provided in section \ref{sec:INNER_MAX}.

\vspace{1mm}
\begin{theorem} \label{LEMMA:INNER_MAX}
    For the binary classification problem, the inner maximum of (\ref{HINGE_ADV_D_minimax}) is attained at $\Delta_\text{hinge}^\star = - \epsilon \cdot \sgn \Big( \sum\iPhat D_i y (v_i - w_i)^\top \Big)$, and the bi-level optimization problem (\ref{HINGE_ADV_D_minimax}) is equivalent to the classic optimization:
    \vspace{-1mm}
    \begin{align}\label{HINGE_ADV_D}
        \min_{(v_i, w_i)\iPhat} & 
        \begin{pmatrix}
            \displaystyle \frac{1}{n} \sum_{k=1}^n \bigg( 1 - y_k \sum\iPhat d_{ik} x_k^\top (v_i - w_i) + \epsilon \biggnorm{\sum\iPhat d_{ik} (v_i - w_i)}_1 \bigg)_+ \\[3.7mm]
            \hfill + \beta \sum\iPhat \big( \norm{v_i}_2 + \norm{w_i}_2 \big)
        \end{pmatrix} \\[.1mm]
        \ST \quad & \ (2 D_i - I_n) X v_i \geq \epsilon \norm{v_i}_1, \;\;
        (2 D_i - I_n) X w_i \geq \epsilon \norm{w_i}_1, \;\; \alliinPhat \nonumber
    \end{align}
    where $d_{ik}$ denotes the $k^\text{th}$ diagonal element of $D_i$.
\end{theorem}

The problem (\ref{HINGE_ADV_D}) is a finite-dimensional convex program that provides an upper bound on (\ref{hinge_adv}), which can be considered as the robust counterpart of (\ref{hinge_std}). We can thus solve (\ref{HINGE_ADV_D}) to robustly train the neural network.
The $\ell_1$ norm term in (\ref{HINGE_ADV_D}) explains the regularization effect of adversarial training.

\subsection{Practical algorithm for convex adversarial training} 

In light of Theorem \ref{thm:prac}, similar to the strategy rendered in Alg \ref{alg:train}, we use a subset of $D_i$ matrices for practical adversarial training. 
For adversarial training, since the $D_i$ matrices depend on the perturbation $\Delta$, we also add randomness to the data matrix $X$ in the sampling process to cover all $D_i$ matrices, leading to Algorithm \ref{alg:adv_train}. By the analogy between Alg \ref{alg:train} and Alg \ref{alg:adv_train}, we expect that even when the cardinality of the subset, denoted as $P_s$, is significantly less than $\widehat{P}$, Alg \ref{alg:adv_train} still provides a good approximation to (\ref{HINGE_ADV_D}). $P_a$ and $S$ are preset parameters that determine the number of random weight samples, with $P_a \cdot S \geq P_s$.

\begin{algorithm}
    \begin{algorithmic}[1]
        \STATE Generate $P_s$ distinct diagonal matrices via the following iterations:
        \FOR{$i=1$ to $P_a$}
            \STATE $a_i \sim \gN (0, I_d)$ i.i.d.
            \STATE $D_{i1} \leftarrow \diag ([X a_i \geq 0])$
            \FOR{$j=2$ to $S$}
                \STATE $R_{ij} \leftarrow [r_1, \dots, r_d]$, where $r_h \sim \gN(\mathbf{0}, I_n), \forall h\in[d]$
                \STATE $D_{ij} \leftarrow \diag ([\overline{X}_{ij} a_i \geq 0])$, where $\overline{X}_{ij} \leftarrow X + \epsilon \cdot \sgn(R_{ij})$
            \ENDFOR
        \ENDFOR
        \STATE Solve \vspace{-4mm}
        \begin{align}\label{hinge_adv_s}
            \min_{(v_i, w_i)\iPs} & 
            \begin{pmatrix}
                \displaystyle \frac{1}{n} \sum_{k=1}^n \bigg( 1 - y_k \sum\iPs d_{ik} x_k^\top (v_i - w_i) + \epsilon \Big|\Big| \sum\iPs d_{ik} (v_i - w_i) \Big|\Big|_1 \bigg)_+ \\
                \hfill + \beta \sum\iPs \big( \norm{v_i}_2 + \norm{w_i}_2 \big) \hfill
            \end{pmatrix} \\[.7mm]
            \ST \quad & \; (2 D_i - I_n) X v_i \geq \epsilon \norm{v_i}_1, \;\; 
            (2 D_i - I_n) X w_i \geq \epsilon \norm{w_i}_1, \;\; \forall i \in [P_s]. \nonumber
        \end{align}
        \vspace{-4mm}
        \STATE \text{Recover $u_1, \ldots, u_{m_s}$ and $\alpha_1, \ldots, \alpha_{m_s}$ from the solution $(v_{\text{rob}s_i}^\star,w_{\text{rob}s_i}^\star)_{i=1}^{P_s}$ of (\ref{hinge_adv_s}) using (\ref{recover_weights})}.
    \end{algorithmic}
    \caption{Practical adversarial training}
    \label{alg:adv_train}
\end{algorithm}

\section{Convex Squared Loss Adversarial Training} \label{sec:sqr_loss}
Squared loss $\ell(\widehat{y}, y) = \frac{1}{2} \enorm{\widehat{y}-y}^2$ is another commonly used loss function in machine learning. It is widely used for regression tasks, but can also be used for classification.

Consider the non-convex training problem of a one-hidden-layer ReLU neural network trained with the $\ell_2$-regularized squared loss:
\begin{align}\label{nonconvex_sl}
    \min_{(u_j, \alpha_j)\jm} \frac{1}{2} \bigg|\bigg| \sum\jm (X u_j)_+ \alpha_j - y & \bigg|\bigg|_2^2 + \frac{\beta}{2}\sum\jm \big( \norm{u_j}_2^2+\alpha_j^2 \big).
\end{align}

Coupling this nominal problem with the uncertainty set $\gX$ gives us the robust counterpart of (\ref{nonconvex_sl}) as
\begin{align} \label{robustnonconvex_sl}
    & \min_{(u_j, \alpha_j)\jm} 
    \begin{pmatrix}
        \displaystyle \max\allDelta \frac{1}{2} \bigg|\bigg| \sum\jm \big( (X+\Delta)u_j \big)_+ \alpha_j - y \bigg|\bigg|_2^2 + \frac{\beta}{2} \sum\jm \big( \norm{u_j}_2^2+\alpha_j^2 \big)
    \end{pmatrix}.
\end{align}

Applying Theorem \ref{THM:CVX_MINIMAX} and Corollary \ref{CORO:ROB_CONSTRAINT} leads to the following formulation as an upper bound on (\ref{robustnonconvex_sl}):
\begin{align}\label{sl_minimax_convex}
    \min_{(v_i, w_i)\iPhat} & 
    \begin{pmatrix}
        \displaystyle \max\allDelta \frac{1}{2} \Bigg|\Bigg|\sum_{i=1}^{\widehat{P}} D_i(X+\Delta)(v_i-w_i) - y \Bigg|\Bigg|_2^2 + \beta \sum_{i=1}^{\widehat{P}} \big( \norm{v_i}_2 + \norm{w_i}_2 \big) \hfill
    \end{pmatrix} \\[-.5mm]
    \ST \quad & (2 D_i - I_n) X v_i \geq \epsilon \norm{v_i}_1, \quad (2 D_i - I_n) X w_i \geq \epsilon \norm{w_i}_1, \quad \alliinPhat. \nonumber
\end{align}

\vspace{-1mm}
Solving the maximization over $\Delta$ in closed form leads to the next result, with the proof provided in Appendix \ref{sec:ROBUST_SOCP}.

\begin{theorem} \label{THEOREM:ROBUST_SOCP}
    The optimization problem (\ref{sl_minimax_convex}) is equivalent to the convex program:
    \vspace{-1.5mm}
    \begin{align}\label{SOCP3}
        & \min_{(v_i, w_i, b_i, c_i)\iPhat, a, z} a + \beta\sum\iPhat (b_i+c_i) \\
        &\ST \ \  (2 D_i - I_n) X v_i \geq \epsilon \norm{v_i}_1, \;\;
        (2 D_i - I_n) X w_i \geq \epsilon \norm{w_i}_1, \;\;
        \norm{v_i}_2\leq b_i, \;\; \norm{w_i}_2\leq c_i, \;\; \alliinPhat \nonumber \\[-1mm]
        &\qquad  z_k\geq \bigg| \sum_{i=1}^{\widehat{P}} D_{ik}x_k^T(v_i-w_i) - y_k \bigg| \nonumber + \epsilon \bigg|\bigg| \sum_{i=1}^{\widehat{P}}D_{ik}(v_i-w_i) \bigg|\bigg|_1, \qquad \forall k\in[n] \nonumber \\[-.5mm]
        &\qquad  z_{n+1}\geq \big| 2a-\tfrac{1}{4} \big|, \;\; \norm{z}_2\leq 2a + \tfrac{1}{4} \nonumber.
   \end{align}
\end{theorem}

Problem (\ref{SOCP3}) is a convex optimization that can be used to train robust neural networks. However, directly using (\ref{SOCP3}) for adversarial training can be intractable due to the large number of constraints that arise when we include all $D$ matrices associated with all $\Delta$ such that $X+\Delta \in \gX$. To this end, the approximate training algorithm (Alg \ref{alg:adv_train}) can be used where we sample a subset of the diagonal matrices $D_1, D_2, \dots, D_{P_s}$. As before, the optimality gap can be measured similar to Theorem \ref{thm:prac}.

\section{Numerical Experiments}\label{sec:exp}
In this section, we focus on the experiments with the proposed convex adversarial training method for the hinge loss. Standard training experiment results that supports Theorem \ref{thm:prac} are provided in Appendix \ref{sec:prac_exp} and experiment results with the squared loss convex adversarial training formulation are provided in Appendix \ref{sl_simulation}.

\begin{figure}[t]
    \centering
    \adjustbox{trim={0} {.01\height} {0} {.07\height}, clip}
    {\includegraphics[width=\textwidth]{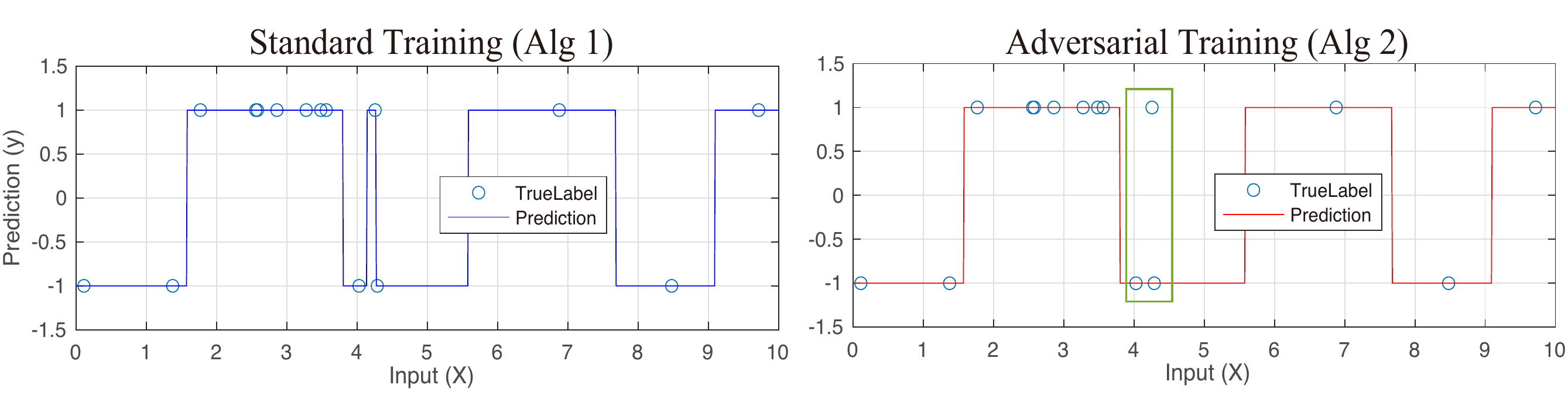}}
    \vspace{-4mm}
    \captionof{figure}{Visualization of binary decision boundaries in 1-dimensional space. The positive point in the highlighted region was considered as an outlier and ignored by Alg \ref{alg:adv_train}.}
    \label{fig:1D-toy}
\end{figure}

To analyze the behavior of the developed problem (\ref{HINGE_ADV_D}) and to visualize the decision boundaries, we first ran Alg \ref{alg:train} and Alg \ref{alg:adv_train} on contrived 1-dimensional data for binary classification. 
A bias term was included by concatenating a column of ones to the data matrix $X$ and the parameter $\epsilon = 0.03$ was used. $X$ included 15 randomly generated points and the labels were randomly generated such that $y \in \{-1, +1\}^{15}$. For all experiments, CVX \citep{cvx} and CVXPY \citep{agrawal2018rewriting,diamond2016cvxpy} with MOSEK \citep{mosek} and SeDuMi \citep{S98guide} solvers were used for solving optimization on a MacBook Pro laptop computer.
Figure \ref{fig:1D-toy} shows that in the 1-dimensional space, Alg \ref{alg:adv_train} ignores the points with conflicting perturbation sets, and ensures all points in the perturbation set to be predicted as the same class.

\begin{figure}
    \centering
    \adjustbox{trim={.01\width} {.01\height} {.01\width} {.01\height},clip}{\includegraphics[width=.45\textwidth]{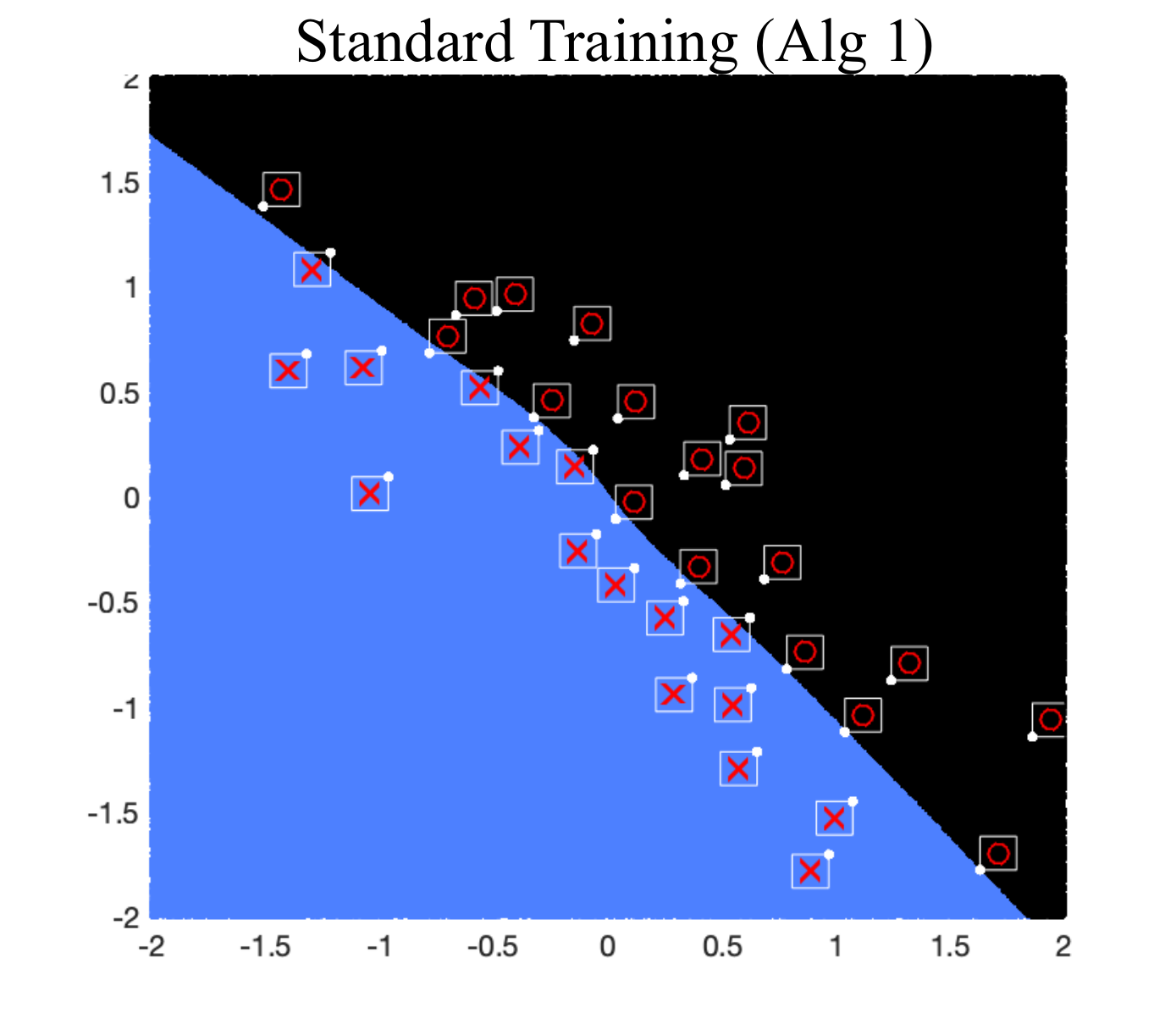}} \adjustbox{trim={.01\width} {.01\height} {.01\width} {.01\height},clip}{\includegraphics[width=.45\textwidth]{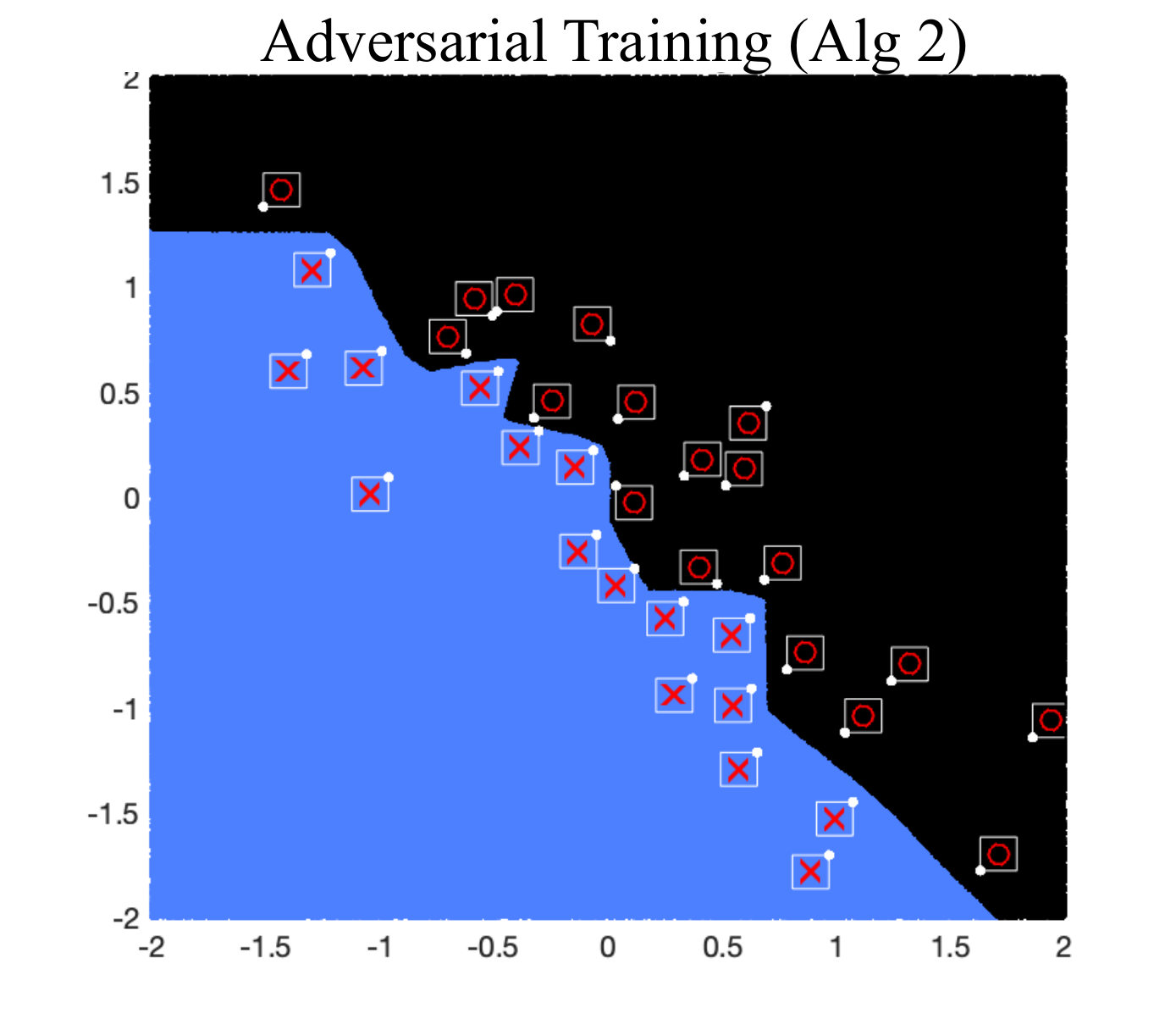}}
    \vspace{-2mm}
    \captionof{figure}{Visualization of binary decision boundaries in 2-dimensional space. The red crosses are positive training points while the red circles are negative points. The region classified as positive is in blue, whereas the negative region is in black. The white box around each training data is the $\ell_\infty$ perturbation bound. The white dot at a vertex of each box is the worst-case perturbation. Alg \ref{alg:adv_train} fitted the perturbation boxes, while the standard training fitted the points.}
    \label{fig:2D-toy}
    \vspace{-2mm}
\end{figure}

Similar experiments were performed on a contrived 2-dimensional dataset. 34 random points were generated in $[-2,2] \times [-2,2]$. The algorithms were run with the parameters $P_s = 360$ and $\epsilon=0.08$ with the bias term again introduced. Figure \ref{fig:2D-toy} shows the decision boundary, confirming that adversarial training (Alg \ref{alg:adv_train}) fits the perturbation boxes as designed.

We then verified the real-world performance of the proposed convex training methods on a subset of the CIFAR-10 image classification dataset \citep{cifar10} for binary classification between the second class and the eighth class. The subset consists of 600 images downsampled to $d=147$. The parameters were $\epsilon=10$, $\beta = 10^{-4}$, and $P_s=36$, corresponding to neural network widths of at most 72.
We used the FGSM and PGD methods to generate adversarial examples and used both clean data and adversarial data to compare the performance of Alg \ref{alg:train}, Alg \ref{alg:adv_train}, the traditional standard training method (standard backprop; abbreviated as GD-std in the tables), and the widely-used adversarial training method: use FGSM or PGD to solve for the inner maximum of (\ref{hinge_adv}) and use gradient descent back-propagation to solve the outer minimization (abbreviated as GD-FGSM and GD-PGD in the tables).
\footnote{We used a modified version of \citep{2019simple} for the implementation of back-propagation algorithms.}

Hinge loss has a flat part that has a zero gradient. To generate adversarial examples even in this part, we treat it as ``leaky hinge loss'' via the model $\max( \zeta (1 - \hat{y} \cdot y), 1 - \hat{y} \cdot y)$, where $\zeta \to 0^+$. Hence, the FGSM calculation (\ref{FGSM}) evaluates to
\vspace{-1mm}
\begin{equation*}\label{FGSM_hinge_mod}
    \textstyle \tilde{x} = x - \epsilon \cdot \sgn \Big( y \cdot \sum_{j:\ x^\top u_j \geq 0} \big( u_j \alpha_j \big) \Big).
\end{equation*}

Similarly, the PGD method (\ref{PGD}) evaluates to
\begin{equation*}\label{PGD_hinge}
    \tilde{x}^{t+1} = \Pi_{\gX} \textstyle \Big( \tilde{x}^t - \gamma \cdot \sgn \big( y \cdot \sum_{j:\ x^\top u_j \geq 0} (u_j \alpha_j) \big) \Big), \quad \tilde{x}^0 = x.
\end{equation*}
where the projection step can be simply performed by clipping the coordinates that deviate more than $\epsilon$ from $x$. In the following experiments, we used $\gamma = \epsilon/30$ and ran PGD for 40 steps.

\newcommand{\pc}{\small{\% }}
\begin{table}[t]
\caption{Average optimal objective and accuracy on clean and adversarial data over 7 runs on the CIFAR-10 database. The numbers in the parentheses are the standard deviations over the 7 runs.}
\label{tbl:cifar}
\begin{center}
\begin{small}
\begin{sc}
\begin{tabular}{lcccl}
\toprule
Method & Clean & FGSM adv. & PGD adv. & Objective \\
\midrule
GD-std                  & 79.56 \pc \tiny{(.4138\%)} & 47.09 \pc \tiny{(.4290\%)} & 45.6 \pc \tiny{(.4796\%)} & .3146 \tiny{(.01101)} \\
GD-FGSM                 & 75.3 \pc \tiny{(3.104\%)} & 61.03 \pc \tiny{(4.763\%)} & 60.99 \pc \tiny{(4.769\%)} & .8370 \tiny{$(6.681\times 10^{-2})$} \\
GD-PGD                  & 76.56 \pc \tiny{(.6038\%)} & 62.48 \pc \tiny{(.2215\%)} & 62.44 \pc \tiny{(.1988\%)} & .8220 \tiny{$(3.933\times 10^{-3})$} \\
Alg \ref{alg:train}     & 81.01 \pc \tiny{(.8090\%)} & .4857 \pc \tiny{(.1842\%)} & .3571 \pc \tiny{(.1239\%)} & $6.910\times 10^{-3}$ \tiny{$(3.020\times 10^{-4})$} \\
Alg \ref{alg:adv_train} & 78.36 \pc \tiny{(.3250\%)} & 66.95 \pc \tiny{(.4564\%)} & 66.81 \pc \tiny{(0.4862\%)} & .6511 \tiny{$(6.903\times 10^{-3})$} \\
\bottomrule
\end{tabular}
\end{sc}
\end{small}
\end{center}
\vspace{-2mm}
\end{table}

The results on the CIFAR-10 dataset are provided in Table \ref{tbl:cifar}. The convex standard training algorithm (Alg \ref{alg:train}) achieved a slightly higher clean accuracy compared with GD-std, and returned a much lower training cost. Such behavior supports the findings of Theorem \ref{thm:prac}. The convex adversarial training algorithm (Alg \ref{alg:adv_train}) achieved better accuracies on clean data and adversarial data compared with GD-FGSM and GD-PGD. While Alg \ref{alg:adv_train} solves the upper-bound problem (\ref{HINGE_ADV_D}), it returned a lower training objective compared with GD-FGSM and GD-PGD, showing that the back-propagation methods failed to find the optimal network. Moreover, the back-propagation methods are very sensitive to initializations and hyperparameter choices. In contrast, since Alg \ref{alg:train} and Alg \ref{alg:adv_train} solve convex programs, they are much less sensitive to initializations and are guaranteed to converge to their global optima.

To further demonstrate the behavior of Alg \ref{alg:train} and Alg \ref{alg:adv_train} and to highlight the improved training efficiency of Alg \ref{alg:adv_train} compared with GD-PGD, we conducted additional experiments on the smaller ``Mammographic Masses'' dataset from the UCI Machine Learning Repository \citep{Dua:2019}. Furthermore, the presence of an $\ell_1$ norm term in the upper-bound formulations (\ref{HINGE_ADV_D}) and (\ref{SOCP3}) indicates that adversarial training with a small $\epsilon$ has a regularizing effect, which can improve generalization, supporting the finding of \citep{DBLP:conf/iclr/KurakinGB17}. Additional experiments on the mammographic masses dataset verify this regularization effect. These experiment results are presented in Appendix \ref{sec:mm_exp}. In all experiments, Alg \ref{alg:adv_train} vastly outperforms Alg \ref{alg:train} on adversarial data, highlighting the contribution of Alg \ref{alg:adv_train}: a novel efficient convex adversarial training procedure that reliably trains robust neural networks. Compared with Alg \ref{alg:train}, Alg \ref{alg:adv_train} retains the advantage in the absence of spurious local minima while achieveing adversarial robustness.

\section{Conclusion} \label{sec:conclu}

In this work, we addressed the following two problems regarding the training of one-hidden-layer fully-connected scalar-output ReLU neural networks:

First, the complexities of prior methods for globally optimizing neural networks are prohibitively high, whereas local search methods often fail to find the global minimum of the cost function. We addressed this issue by proposing a stochastic approximation procedure to form convex training programs. This procedure provably improves the training efficiency exponentially compared with the exact counterpart and outperforms the back-propagation method. 

Second, prior adversarial training methods struggle to find optimal robust neural networks due to optimization difficulties. We addressed this issue by deriving a new adversarial training formulation and showed that this formulation is indeed convex and efficiently solvable for the case of hinge loss and squared losses. To the best of our knowledge, this is the first convex program for the adversarial training problem. The structure of the proposed convex formulations explains the regularizing effect of adversarial training. Through numerical experiments on various datasets, we demonstrated that the proposed method fits the perturbation boxes around the training data points as designed and noticeably improves the performance on adversarial test data compared with previous methods.

\bibliographystyle{plainnat}
\bibliography{paper}

\newpage
\appendix

\section{Appendix}

\subsection{Additional experiments}
\subsubsection{Practical standard training experiments} \label{sec:prac_exp}
\begin{figure*}
    \centering
    \includegraphics[width=0.49\textwidth]{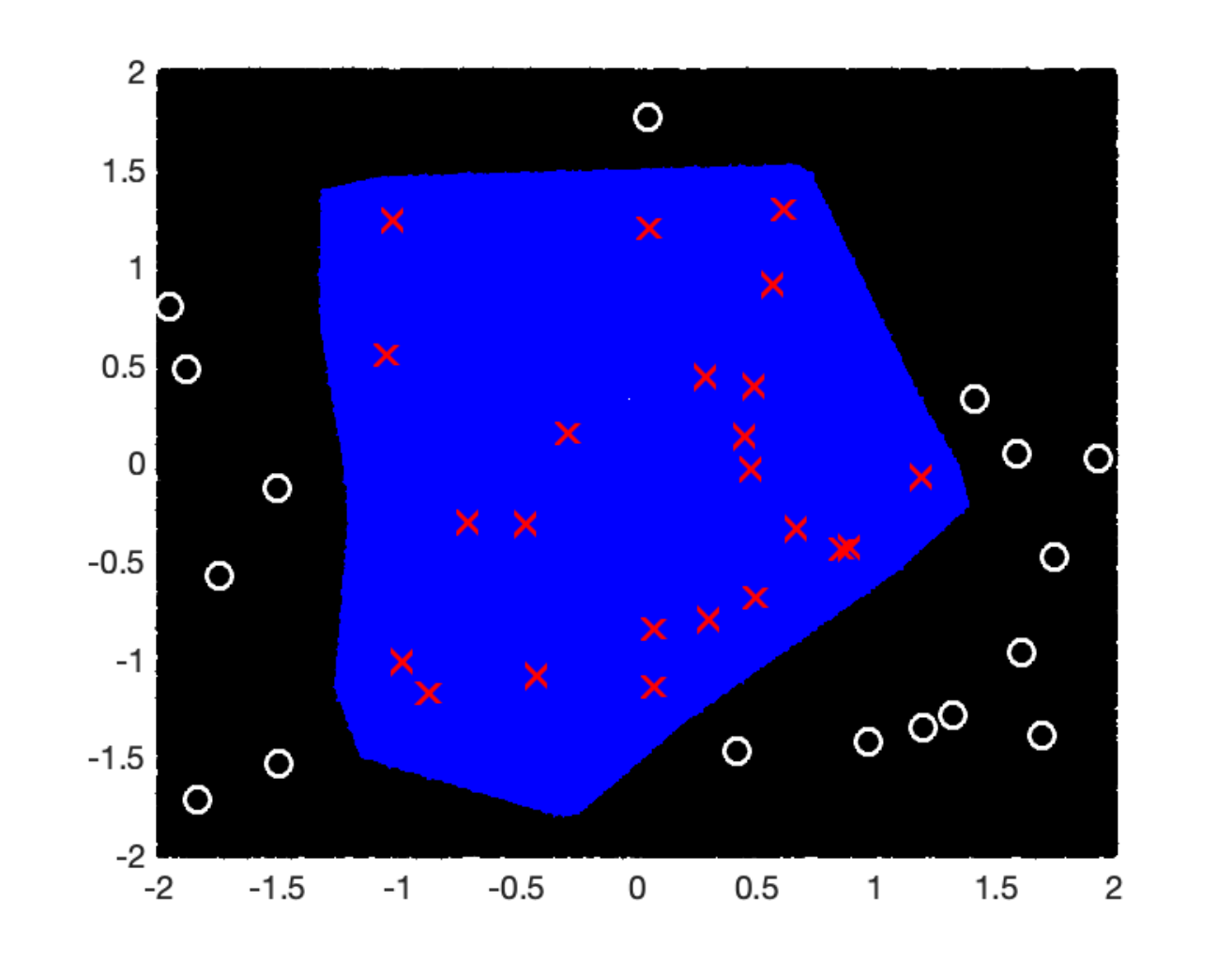} \includegraphics[width=0.49\textwidth]{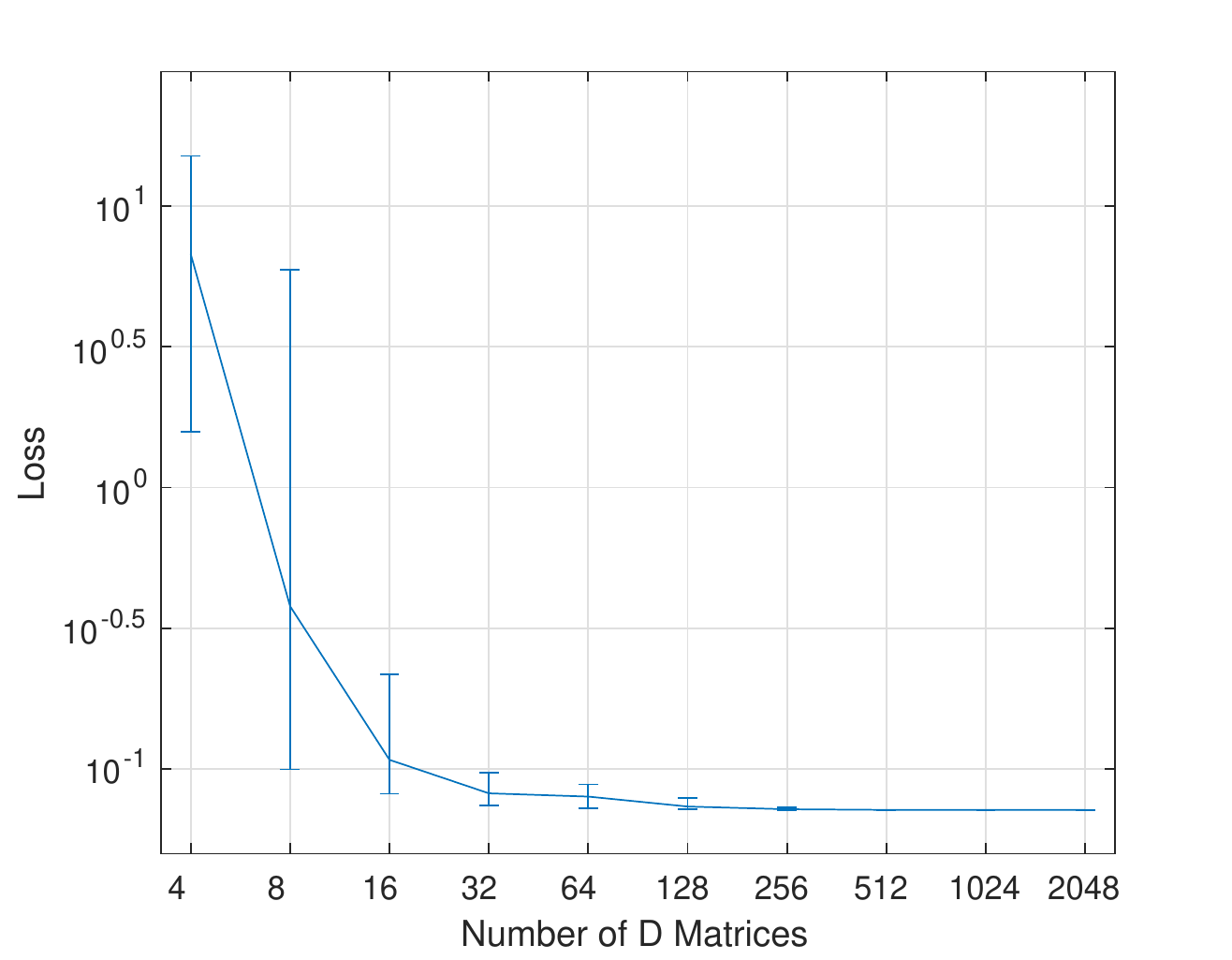}
    \vspace{-.5mm}
    \caption{The left figure is a randomized 2-dimensional dataset. The right figure is the optimized training loss for each $P_s$. The red crosses are positive training points and the white circles are negative training points. The region classified as positive is in blue, whereas the negative region is in black. When $P_s$ reaches 128, the mean and variance of the optimized loss become very small.}
    \label{fig:stability}
\end{figure*}

In this part of the appendix, we use numerical experiments to demonstrate the efficacy of the practical training algorithm (Alg \ref{alg:train}) and to show the level of suboptimality of the neural network trained using Alg \ref{alg:train}.
The experiment was performed on a randomly-generated dataset with $n=40$ and $d=2$. The upper bound of the number of ReLU activation patterns is $4 \big( \frac{e(39)}{2} \big)^2 = 11239$. We ran Alg \ref{alg:train} to train neural networks using the hinge loss with the number of $D_i$ matrices equal to $4, 8, 16, \dots, 2048$ \footnote{$P_s$ was set to 81920, and the sampling was terminated when a sufficient number of $D_i$ matrices was generated. $\beta$ was chosen as $10^{-4}$.} and compared the optimized loss. We repeated this experiment 15 times for each setting, and plotted the loss in Figure \ref{fig:stability}. The error bars show the loss values achieved in the best and the worst runs. When there are more than 128 matrices (much less than the theoretical bound on $P$), Alg \ref{alg:train} yields consistent and favorable results. Further increasing the number of $D$ matrices does not produce a significantly lower loss. By Theorem \ref{thm:prac}, $P_s = 128$ corresponds to $\psi \xi = 0.318$.

\subsubsection{Experiments on the ``Mammographic Masses'' dataset} \label{sec:mm_exp}
In this part of the appendix, we use experiments on the smaller ``Mammographic Masses'' dataset to further demonstrate the behavior of the networks trained with Alg \ref{alg:train} and Alg \ref{alg:adv_train}, as well as the superior efficiency of the algorithms. Furthermore, we verify that when $\epsilon$ is small, adversarial training has a regularization effect.

We removed instances containing NaNs and randomly selected 70\% of the data for training and 30\% for testing, resulting in $n=581$ and $d=5$. The hyperparameters were $\epsilon = 0.12$, $\beta = 10^{-4}$, and $P_s = 120$, corresponding to neural network widths of at most 240. $z$-score standardization was performed before the neural networks were trained with the hinge loss via various methods. The results are shown in Table \ref{tbl:accuracies}.

\begin{table}[t]
\caption{Average optimal objective, CPU time, and prediction accuracy on clean and adversarial data over 20 runs on the Mammographic Masses dataset.}
\label{tbl:accuracies}
\begin{center}
\begin{small}
\begin{sc}
\begin{tabular}{lccccc}
\toprule
Method & Clean & FGSM adv. & PGD adv. & Objective & CPU Time (s) \\
\midrule
GD-std              & 81.14 \% & 57.83 \% & 52.75 \% & .4144 & 4.996 \\
GD-PGD                   & 81.35 \% & 75.82 \% & 74.00 \% & .6955 & 143.0 \\
Alg \ref{alg:train}      & 79.80 \% & 43.41 \% & 31.14 \% & .2428 & 36.92 \\
Alg \ref{alg:adv_train}  & 80.00 \% & 75.72 \% & 75.68 \% & .9663 & 38.26 \\
\bottomrule
\end{tabular}
\end{sc}
\end{small}
\end{center}
\vspace{-2mm}
\end{table}

As expected, Alg \ref{alg:train} returned a lower training cost than GD-std. However, such advantage in optimization was not reflected in the prediction accuracy. Alg \ref{alg:adv_train} returned a higher objective than GD-PGD did, which is expected since Alg \ref{alg:adv_train} solves the upper-bound problem. From a computational burden standpoint, GD-PGD can be slow due to the iterative process of generating adversarial training examples, whereas Alg \ref{alg:adv_train} was noticeably faster. 
In terms of the prediction accuracy, Alg \ref{alg:adv_train} slightly outperformed GD-PGD on PGD adversaries. 
The explanation for the slightly lower accuracies of Alg \ref{alg:adv_train} on clean and FGSM data is that when the adversarial perturbation boxes of the training data points overlap, there is a trade-off between ``assigning the same class to a local neighborhood'' and ``assigning the correct class to the most part of the neighborhood''. From experiments, we have observed that Alg \ref{alg:adv_train} prioritized the former, emphasizing robustness over clean accuracies. Figure \ref{fig:1D-toy} illustrates this property with a 1-dimensional example.

To empirically verify the regularization effect of adversarial training, we chose a small adversary strength ($\epsilon=0.005$), and ran both algorithms 15 times with the number of $D_i$ matrices $P_s$ equal to 80. The average accuracy on clean data was 82.97\% for Alg \ref{alg:adv_train} and 79.44\% for Alg \ref{alg:train}. This result verifies that adversarial training with a small $\epsilon$ can be used as a regularizer to alleviate overfitting and thereby improve accuracy even on clean test data, as suggested in \citep{DBLP:conf/iclr/KurakinGB17}.

\subsubsection{Experiments with squared loss adversarial training} \label{sl_simulation}
\begin{figure}[t]
    \centering
    \begin{minipage}{.45\textwidth}
      \centering
      \includegraphics[width=\textwidth]{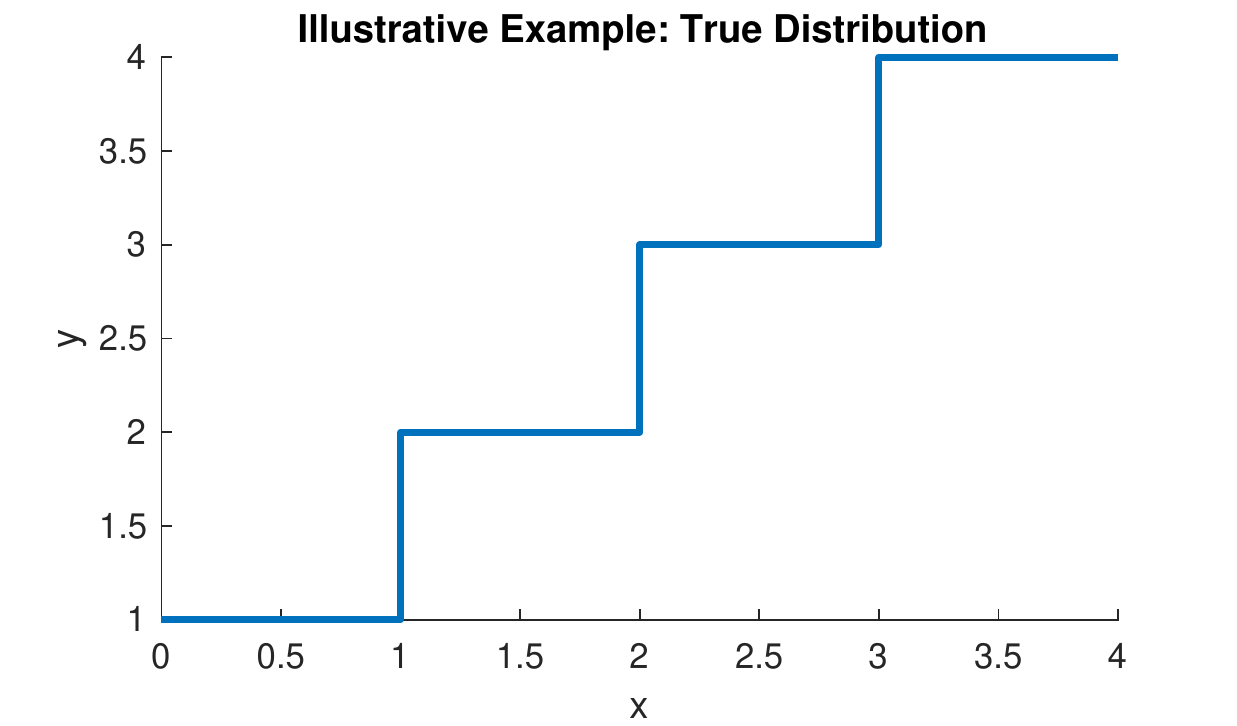}
      \captionof{figure}{True relationship between data ($x$) and target ($y$) used in the illustrative example in Section \ref{sl_simulation}. Training (with $n=8$ points) and test (with $n=100$ points) sets are uniformly sampled from the distribution.}
      \label{fig:te}
    \end{minipage}
    \quad
    \begin{minipage}{.51\textwidth}
      \centering
      \includegraphics[width=\textwidth]{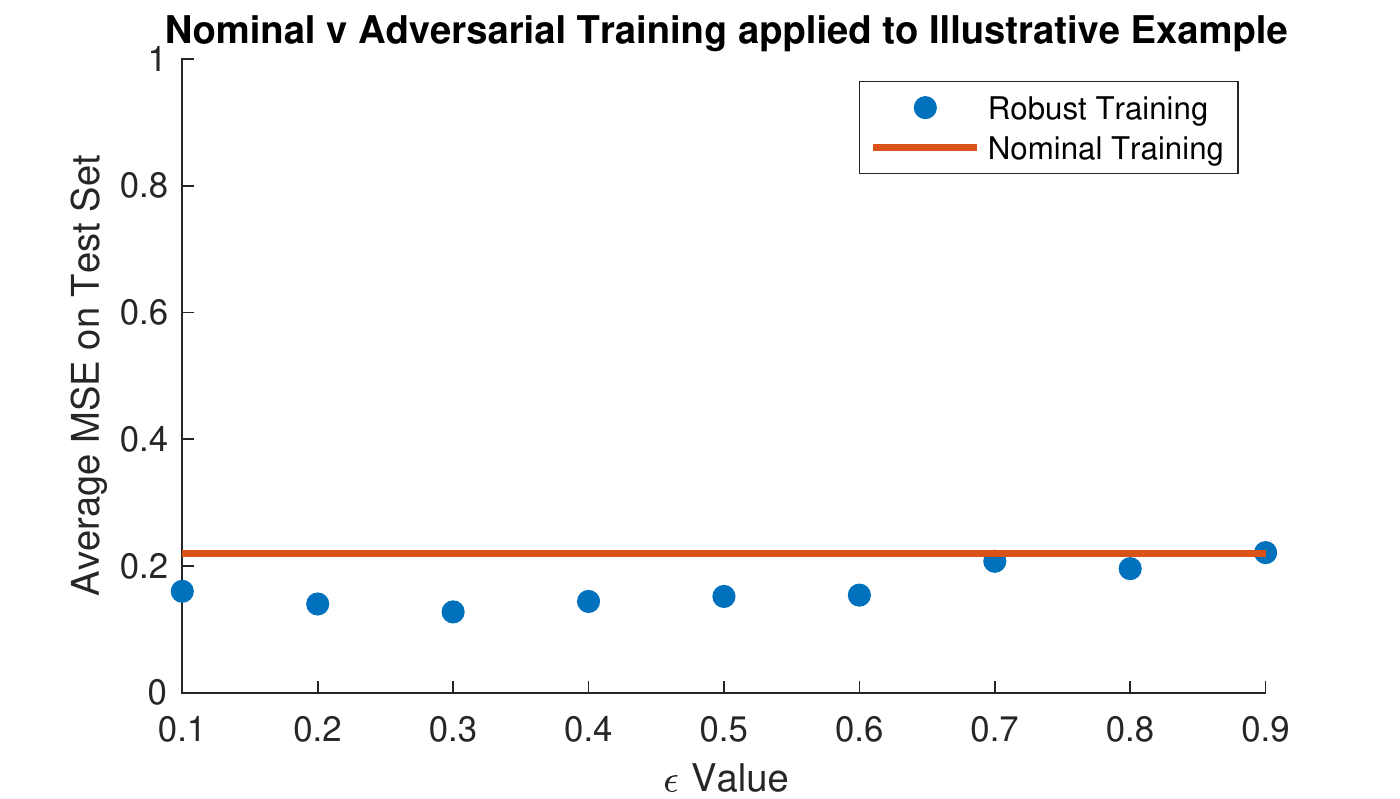}
      \captionof{figure}{This plot shows that the robust training approach (\ref{SOCP3}) outperforms the standard approach for different $\eps\in\{0.1, ..., 0.9\}$ on the dataset studied in Section \ref{sl_simulation}.}
      \label{fig:exp_sl}
    \end{minipage}
    \vspace{-2mm}
\end{figure}

In this part of the appendix, the performance of the developed problem (\ref{SOCP3}) was compared with the standard training problem (\ref{convex_general}) on a contrived 1-dimensional dataset.
Figure \ref{fig:te} shows the true relationship between the data vector $X$ and the target output $y$. Throughout this experiment, training data were constructed by uniformly sampling 8 points from this distribution and test data were similarly constructed by uniformly sampling 100 points.
A bias term was included by concatenating a column of ones to $X$.

The training and testing procedure was repeated for 100 trials with standard training (Alg \ref{alg:train}). For the adversarial training (Alg \ref{alg:adv_train}), we varied the perturbation radius $\epsilon = 0.1, \dots, 0.9$. The training and testing procedure was carried out for 10 trials for each $\epsilon$. Figure \ref{fig:exp_sl} reports the average test mean square error (MSE) for each setup.

The adversarial training procedure outperforms standard training for all $\epsilon$ choices. We further observe that the average MSE is the lowest at $\epsilon\approx 0.3$. This behavior arises as the robust problem attempts to account for all points within the uncertainty interval around the sampled training points. When $\epsilon$ is too small, the robust problem approaches the standard training problem. Larger values of $\epsilon$ cause the uncertainty interval to overestimate the constant regions of the true distribution, increasing the MSE.

\subsection{Proof of Theorem \ref{thm:prac}} \label{sec:pracproof}
We start by recasting the constraint of (\ref{eq:dual}) as $\max_{\norm{u}_2\leq1} |v^\top (Xu)_+| \leq \beta$, and obtain
\begin{align*}
    \max_{\norm{u}_2\leq1} \big| v^\top (Xu)_+ \big| = & \max_{\norm{u}_2\leq1} \big| v^\top \diag([Xu\geq0]) X u \big| = \max_{i\in[P]} \bigg( \begin{matrix} \displaystyle \max_{\substack{\norm{u}_2 \leq 1 \\ (2D_i - I_n) Xu \geq 0}} \end{matrix} \big| v^\top D_i X u \big| \bigg),
\end{align*}
where the last equality holds by the definition of the $D_i$ matrices: $D_1 \dots, D_P$ are all distinct matrices that can be formed by $\diag([Xu\geq0])$ for some $u\in\R^d$. The constraint $(2D_i - I_n) X u \geq 0$ is equivalent to $D_i X u \geq 0$ and $(I_n - D_i) X u \leq 0$, which forces $D_i = \diag([Xu\geq0])$ to hold.

Therefore, (\ref{eq:dual}) can be recast as
\begin{align} \label{eq:dual2}
    \max_v -\ell^*(v) \qquad \ST \; \begin{matrix} \displaystyle \max_{\substack{\norm{u}_2 \leq 1 \\ (2D_i - I_n) Xu \geq 0}} \end{matrix} \big| v^\top D_i X u \big| \leq \beta,\ \forall i \in [P].
\end{align}

To form a tractable convex program that provides an approximation to (\ref{eq:dual2}), one can independently sample a subset of the diagonal matrices. One possible sampling procedure is presented in Alg \ref{alg:train}. The sampled matrices, denoted as $D_1, \dots, D_{P_s}$, can be used to construct the relaxed problem:
\begin{align} \label{eq:drelax}
    d_{s1}^\star = \max_v -\ell^*(v) \qquad \ST \; \begin{matrix} \displaystyle \max_{\substack{\norm{u}_2 \leq 1 \\ (2D - I_n) Xu \geq 0}} \end{matrix} \big| v^\top D_i X u \big| \leq \beta,\ \alliinPs.
\end{align}

The optimization problem (\ref{eq:drelax}) is convex with respect to $v$. \citep{pmlr-v119-pilanci20a} has shown that (\ref{eq:dual2}) has the same optimal objective as its dual problem (\ref{convex_general}). By following precisely the same derivation, it can be shown that (\ref{eq:drelax}) has the same optimal objective as (\ref{eq:prac_clean}) and $p_{s1}^\star = d_{s1}^\star$. Moreover, if an additional diagonal matrix $D_{P_s+1}$ is independently randomly sampled to form (\ref{eq:prac_clean2}), then we also have $p_{s2}^\star = d_{s2}^\star$, where
\begin{align*} \label{eq:drelax2}
    d_{s2}^\star = \max_v -\ell^*(v) \qquad \ST \; \begin{matrix} \displaystyle \max_{\substack{\norm{u}_2 \leq 1 \\ (2D - I_n) X u \geq 0}} \end{matrix} \big| v^\top D_i X u \big| \leq \beta,\ \alliinPss.
\end{align*}

Thus, the level of suboptimality of (\ref{eq:drelax}) compared with (\ref{eq:dual2}) is the level of suboptimality of (\ref{eq:prac_clean}) compared with (\ref{convex_general}). It can be concluded from \citep[Theorem 1]{Calafiore2005} that if $P_s \geq \frac{n+1}{\psi \xi} - 1$, then with probability no smaller than $1-\xi$, the solution $v^\star$ to the sampled convex problem (\ref{eq:drelax}) satisfies
\begin{equation*}
    \sP \Big\{ D\in\gD \ : \begin{matrix} \displaystyle \max_{\substack{\norm{u}_2 \leq 1 \\ (2D - I_n) Xu \geq 0}} \end{matrix} \big| v^{\star\top} D X u \big| > \beta \Big\} \leq \psi.
\end{equation*}
where $\gD$ denotes the set of all diagonal matrices that can be formed by $\diag([X u \geq 0])$ for some $u\in\R^d$, which is the set formed by $D_1, \dots, D_P$.

Since $D_{P_s+1}$ is randomly sampled from $\gD$, we have
\begin{equation*}
    \sP \Big\{ D\in\gD \ : \begin{matrix} \displaystyle \max_{\substack{\norm{u}_2 \leq 1 \\ (2D - I_n) Xu \geq 0}} \end{matrix} \big| v^{\star\top} D X u \big| > \beta \Big\} =
    \sP \Big\{ \begin{matrix} \displaystyle \max_{\substack{\norm{u}_2 \leq 1 \\ (2D_{P_s+1} - I_n) X u \geq 0}} \end{matrix} \big| v^{\star\top} D_{P_s+1} X u \big| > \beta \Big\}
\end{equation*}

Thus, with probability no smaller than $1-\xi$,
\begin{equation*}
    \sP \Big\{ \begin{matrix} \displaystyle \max_{\substack{\norm{u}_2 \leq 1 \\ (2D_{P_s+1} - I_n) X u \geq 0}} \end{matrix} \big| v^{\star\top} D_{P_s+1} X u \big| > \beta \Big\} \leq \psi.
\end{equation*}

Moreover, $d_{s2}^\star < d_{s1}^\star$ if and only if $\big| v^{\star\top} D_{P_s+1} X u \big| > \beta$ with $d_{s2}^\star = d_{s1}^\star$ otherwise. The proof is completed by noting that $p_{s1}^\star = d_{s1}^\star$ and $p_{s2}^\star = d_{s2}^\star$. $\hfill \blacksquare$

\subsection{Proof of Theorem \ref{THM:CVX_MINIMAX}} \label{sec:CVX_MINIMAX}
Before proceeding with the proof, we first present the following result borrowed from \citep{pmlr-v119-pilanci20a}.

\begin{lemma} \label{LEMMA:RECOVER}
    For a given data matrix $X$ and $(v_i, w_i)\iP$, if $(2 D_i - I_n) X v_i \geq 0$ and $(2 D_i - I_n) X w_i \geq 0$ for all $i \in [P]$, then we can recover the corresponding neural network weights $(u_{v,w_j}, \alpha_{v,w_j})\jms$ using the formulas in (\ref{recover_weights}), and
    \vspace{-1mm}
    \begin{align} \label{MertEquality}
    & \ell \bigg( \sum\iP D_i X (v_i - w_i), y \bigg) + \beta \sum\iP \big( \norm{v_i}_2 + \norm{w_i}_2 \big) \nonumber \\
    = & \ell \bigg( \sum\jms (X u_{v,w_j})_+ \alpha_{v,w_j}, y \bigg) + \frac{\beta}{2} \sum\jms \big( \norm{u_{v,w_j}}_2^2 + \alpha_{v,w_j}^2 \big).
    \end{align}
\end{lemma}

Theorem \ref{THM:PILANCI} implies that (\ref{nonconvex_general}) has the same objective value as the following finite-dimensional convex optimization problem:
\begin{equation}\label{theorem_5}
\begin{aligned}
    q^\star = \min_{(v_i, w_i)\iP} & \ell \bigg( \sum\iP D_i X (v_i - w_i), y \bigg) + \beta \sum\iP \big(\norm{v_i}_2 + \norm{w_i}_2\big) \\
    \ST \quad & (2 D_i - I_n) X v_i \geq 0, \; (2 D_i - I_n) X w_i \geq 0, \;\; \alliinP
\end{aligned}
\end{equation}
where $D_1, \dots, D_P$ are all of the matrices in the set of matrices $\gD$, which is defined as the set of all distinct diagonal matrices $\diag([X u \geq 0])$ that can be obtained for all possible $u \in \mathbb{R}^d$. We recall that the optimal neural network weights can be recovered using (\ref{recover_weights}).

Consider the optimization problem (\ref{redundant_D})
\begin{equation}\label{redundant_D}
\begin{aligned}
    \widetilde{q}^\star = \min_{(v_i, w_i)\itildeP} & \ell \bigg( \sum\itildeP D_i X (v_i - w_i), y \bigg) + \beta \sum\itildeP \big(\norm{v_i}_2 + \norm{w_i}_2\big) \\
    \ST \quad & (2 D_i - I_n) X v_i \geq 0, \; (2 D_i - I_n) X w_i \geq 0, \;\; \forall i \in [\widetilde{P}]
\end{aligned}
\end{equation}
where additional $D$ matrices, denoted as $D_{P+1}, \dots, D_{\widetilde{P}}$, are introduced. These additional matrices are still diagonal with each entry being either 0 or 1, while they do not belong to $\gD$. They represent ``infeasible hyperplanes'' that cannot be achieved by the sign pattern of $X u$ for any $u\in\R^d$. 

\begin{lemma}\label{LEMMA:RED_D}
    It holds that $\widetilde{q}^\star = q^\star$, meaning that the optimization problem (\ref{redundant_D}) has the same optimal objective as (\ref{theorem_5}).
\end{lemma}
The proof of Lemma \ref{LEMMA:RED_D} is given in section \ref{sec:RED_D}.

The robust minimax training problem (\ref{robust_general}) considers an uncertain data matrix $X+\Delta$. Different values of $X+\Delta$ within the uncertainty set $\gU$ can result in different $D$ matrices. 
Now, we define $\widehat{\gD} = \bigcup_\Delta \gD_\Delta$, where $\gD_\Delta$ is the set of diagonal matrices for a particular $\Delta$ such that $X+\Delta \in \gU$.
By construction, we have $\gD_\Delta \subseteq \widehat{\gD}$ for every $\Delta$ such that $X+\Delta \in \gU$. Thus, if we define $D_1, \dots, D_{\widehat{P}}$ as all matrices in $\widehat{\gD}$, then for every $\Delta$ with the property $X+\Delta \in \gU$, the optimization problem
\begin{equation} \label{cvx_allD}
\begin{aligned}
    \min_{(v_i, w_i)\iPhat} & \ell \bigg( \sum\iPhat D_i (X+\Delta) (v_i - w_i), y \bigg) + \beta \sum\iPhat (\norm{v_i}_2 + \norm{w_i}_2) \\
    \ST \quad & (2 D_i - I_n) (X+\Delta) v_i \geq 0, \; (2 D_i - I_n) (X+\Delta) w_i \geq 0, \;\: \alliinPhat
\end{aligned}
\end{equation}
is equivalent to
\begin{equation*}
        \min_{(u_j, \alpha_j)\jm} \ell \bigg( \sum\jm ((X+\Delta)u_j)_+ \alpha_j, y \bigg) + \frac{\beta}{2} \sum\jm \big( \norm{u_j}_2^2 + \alpha_j^2 \big)
\end{equation*}
as long as $m \geq \widehat{m}^\star$ with $\widehat{m}^\star = |\{i : v_i^\star(\Delta) \neq 0\}| + |\{i : w_i^\star(\Delta) \neq 0\}|$, where $(v_i^\star(\Delta), w_i^\star(\Delta))\iPhat$ denotes an optimal point to (\ref{cvx_allD}).

Now, we focus on the minimax training problem with a convex objective given by
\begin{equation} \label{minimax1}
\begin{aligned}
    \min_{(v_i, w_i)\iPhat\in\gF} 
    \begin{pmatrix}
        \displaystyle \max\allDeltaU \ell \bigg( \sum\iPhat D_i (X+\Delta) (v_i - w_i), y \bigg) + \beta \sum\iPhat \big( \norm{v_i}_2 + \norm{w_i}_2 \big) \hfill \\[4mm]
        \quad\ \ST \;\; (2 D_i - I_n) (X+\Delta) v_i \geq 0, \; (2 D_i - I_n) (X+\Delta) w_i \geq 0, \; \alliinPhat
    \end{pmatrix},
\end{aligned}
\end{equation}
where $\gF$ is defined as:
\vspace{1mm}
\begin{equation*}
    \bigg\{ (v_i, w_i)\iPhat \; \bigg| \; \begin{matrix}
        \; \exists\Delta:X+\Delta\in\gU \hfill \\ \ST \; (2 D_i - I_n) (X+\Delta) v_i \geq 0, \; (2 D_i - I_n) (X+\Delta) w_i \geq 0, \; \alliinPhat
    \end{matrix} \bigg\}.
\end{equation*}

The introduction of the feasible set $\gF$ is to avoid the situation where the inner maximization over $\Delta$ is infeasible and the objective becomes $- \infty$, leaving the outer minimization problem unbounded.

\newcommand{\DeltaStar}{\Delta^\star_{v,w}}
\newcommand{\DeltaStarTil}{\widetilde{\Delta}^\star_{v,w}}
\newcommand{\DeltaSStar}{\Delta^{\star\star}_{v,w}}

Moreover, consider the following problem:
\begin{equation} \label{minimax2}
\begin{aligned}
    \min_{(v_i, w_i)\iPhat} & 
    \begin{pmatrix}
        \displaystyle \ell \bigg( \sum\iPhat D_i (X+\DeltaStar) (v_i - w_i), y \bigg) + \beta \sum\iPhat \big( \norm{v_i}_2 + \norm{w_i}_2 \big)
    \end{pmatrix} \\
    \ST \quad & (2 D_i - I_n) (X+\DeltaStar) v_i \geq 0, \; (2 D_i - I_n) (X+\DeltaStar) w_i \geq 0, \;\; \alliinPhat
\end{aligned}
\end{equation}
where $\DeltaStar$ is the optimal point for $\displaystyle \max\allDeltaU \ell \bigg( \sum\iPhat D_i (X+\Delta) (v_i - w_i), y \bigg)$. 
Note that the inequality constraints are dropped for the maximization here compared to (\ref{minimax1}).

The optimization problem (\ref{minimax1}) gives a lower bound on (\ref{minimax2}). To prove this, we first rewrite (\ref{minimax2}) as:
\begin{align*}
    \min_{(v_i, w_i)\iPhat} & f \big( (v_i, w_i)\iPhat \big) \text{, where } f \big( (v_i, w_i)\iPhat \big) = \\
    & \begin{cases}
        \ell \Big( \sum\iPhat D_i (X+\DeltaStar) (v_i - w_i), y \Big) & (2 D_i - I_n) (X+\DeltaStar) v_i \geq 0, \; \alliinPhat \\[-1mm]
        \hspace{2cm} + \beta \sum\iPhat \big( \norm{v_i}_2 + \norm{w_i}_2 \big), & (2 D_i - I_n) (X+\DeltaStar) w_i \geq 0, \; \alliinPhat \\[2.5mm]
        + \infty, & \text{otherwise}.
    \end{cases}
\end{align*}

Now, we analyze (\ref{minimax1}). Consider three cases:

Case 1: For some $(v_i, w_i)\iPhat$, $\DeltaStar$ is optimal for the inner maximization of (\ref{minimax1}) and the inequality constraints are inactive. This happens whenever $\DeltaStar$ is feasible for the particular choice of $(v_i, w_i)\iPhat$. In other words, $(2 D_i - I_n) (X+\DeltaStar) v_i \geq 0$ and $(2 D_i - I_n) (X+\DeltaStar) w_i \geq 0$ hold true for all $i\in[\widehat{P}]$. For these $(v_i, w_i)\iPhat$, we have:
\begin{align*}
    & \begin{pmatrix}
        \displaystyle \max\allDeltaU \ell \bigg( \sum\iPhat D_i (X+\Delta) (v_i - w_i), y \bigg) + \beta \sum\iPhat \big( \norm{v_i}_2 + \norm{w_i}_2 \big) \\[4mm]
        \ST \:\: (2 D_i - I_n) (X+\Delta) v_i \geq 0, \; (2 D_i - I_n) (X+\Delta) w_i \geq 0, \; \alliinPhat
    \end{pmatrix} \\[1mm]
    & \hspace{15mm} = \ell \bigg( \sum\iPhat D_i (X+\DeltaStar) (v_i - w_i), y \bigg)
        + \beta \sum\iPhat \big( \norm{v_i}_2 + \norm{w_i}_2 \big)
\end{align*}

Case 2: For some $(v_i, w_i)\iPhat$, $\DeltaStar$ is infeasible, while some $\Delta$ within the perturbation bound satisfies the inequality constraints. Suppose that among the feasible $\Delta$'s,
\begin{align*}
    \DeltaStarTil = & \argmax\allDeltaU \ell \bigg( \sum\iPhat D_i (X+\Delta) (v_i - w_i), y \bigg) + \beta \sum\iPhat \big( \norm{v_i}_2 + \norm{w_i}_2 \big) \\
    & \ST \:\: (2 D_i - I_n) (X+\Delta) v_i \geq 0, \; (2 D_i - I_n) (X+\Delta) w_i \geq 0, \; \alliinPhat.
\end{align*}
In this case,
\begin{align*}
    & \begin{pmatrix}
        \displaystyle \max\allDeltaU \ell \bigg( \sum\iPhat D_i (X+\Delta) (v_i - w_i), y \bigg) + \beta \sum\iPhat \big( \norm{v_i}_2 + \norm{w_i}_2 \big) \hfill \\[4mm]
        \ST \:\: (2 D_i - I_n) (X+\Delta) v_i \geq 0, \; (2 D_i - I_n) (X+\Delta) w_i \geq 0, \; \alliinPhat
    \end{pmatrix} \\[1mm]
    & \hspace{15mm} = \ell \bigg( \sum\iPhat D_i (X+\DeltaStarTil) (v_i - w_i), y \bigg)
        + \beta \sum\iPhat \big( \norm{v_i}_2 + \norm{w_i}_2 \big)
\end{align*}

Case 3: For all other $(v_i, w_i)\iPhat$, the objective value is $+ \infty$ since they do not belong to $\gF$.

Therefore, (\ref{minimax1}) can be rewritten as
\begin{align*}
    & \min_{(v_i, w_i)\iPhat} g \big( (v_i, w_i)\iPhat \big), \ \text{where } g \big( (v_i, w_i)\iPhat \big) = \\
    & \; \begin{cases}
        \ell \Big( \sum\iPhat D_i (X+\DeltaStar) (v_i - w_i), y \Big) & (2 D_i - I_n) (X+\DeltaStar) v_i \geq 0, \; \alliinPhat \\[-1mm]
        \hspace{23mm} + \beta \sum\iPhat \big( \norm{v_i}_2 + \norm{w_i}_2 \big), & (2 D_i - I_n) (X+\DeltaStar) w_i \geq 0, \; \alliinPhat \\[2.5mm]
        & \exists j: (2 D_j - I_n) (X+\DeltaStar) v_j < 0 \\
        \ell \Big( \sum\iPhat D_i (X+\DeltaStarTil) (v_i - w_i), y \Big) & \hspace{2.3mm} \text{or} \ (2 D_j - I_n) (X+\DeltaStar) w_j < 0 \\[-1mm]
        \hspace{23mm} + \beta \sum\iPhat \big( \norm{v_i}_2 + \norm{w_i}_2 \big), & \exists \Delta: \ (2 D_i - I_n) (X+\Delta) v_i \geq 0, \; \alliinPhat \\
        & \hspace{8.7mm} (2 D_i - I_n) (X+\Delta) w_i \geq 0, \; \alliinPhat \\[2.5mm]
        + \infty, & \text{otherwise}
    \end{cases}
\end{align*}

Hence, $g((v_i, w_i)\iPhat) = f((v_i, w_i)\iPhat)$ for all $(v_i, w_i)\iPhat$ belonging to the first and the third cases. $g((v_i, w_i)\iPhat) < f((v_i, w_i)\iPhat)$ for all $(v_i, w_i)\iPhat$ belonging to the second case. Thus, $\min_{(v_i, w_i)\iPhat} g((v_i, w_i)\iPhat) \leq \min_{(v_i, w_i)\iPhat} f((v_i, w_i)\iPhat)$. This concludes that (\ref{minimax1}) is a lower bound to (\ref{minimax2}). 

Let $(v_{\text{minimax}_i}^\star, w_{\text{minimax}_i}^\star)\iPhat$ denote an optimal point for (\ref{minimax2}). It is possible that for some $\Delta: X+\Delta \in \gU$, the constraints $(2 D_i - I_n) (X+\Delta) v_{\text{minimax}_i}^\star \geq 0$ and $(2 D_i - I_n) (X+\Delta) w_{\text{minimax}_i}^\star \geq 0$ are not satisfied for all $i\in[\widehat{P}]$.
In light of Lemma \ref{LEMMA:RECOVER}, at those $\Delta$ where such constraints are violated, the convex problem (\ref{minimax2}) does not reflect the cost of the neural network. For these infeasible $\Delta$, the input-label pairs $(X+\Delta, y)$ can have a high cost in the neural network and potentially become the worst-case adversary. However, these $\Delta$ are ignored in (\ref{minimax2}) due to the infeasibility.
Since adversarial training aims to minimize the cost over the worst-case adversaries generated upon the training data whereas (\ref{minimax2}) may sometimes miss the worst-case adversaries, (\ref{minimax2}) does not fully accomplish the task of adversarial training.
In fact, by applying Theorem \ref{THM:PILANCI} and Lemma \ref{LEMMA:RED_D}, it can be verified that (\ref{minimax1}) and (\ref{minimax2}) are lower bounds to (\ref{robust_general}) as long as $m \geq \widehat{m}^\star$:
\begin{equation*}
    \begin{aligned}
        & \min_{(u_j, \alpha_j)\jm} 
        \begin{pmatrix}
            \displaystyle \max\allDeltaU \ell \bigg( \sum_{j=1}^m \big( (X+\Delta) u_j \big)_+ \alpha_j, y \bigg) + \frac\beta2 \sum_{j=1}^m \Big( \norm{u_j}_2^2 + \alpha_j^2 \Big)
        \end{pmatrix} \\
        & \hspace{15mm} \geq \min_{(u_j, \alpha_j)\jm} \ell \bigg( \sum_{j=1}^m \big( (X+\DeltaStar) u_j \big)_+ \alpha_j, y \bigg) + \frac\beta2 \sum_{j=1}^m \Big( \norm{u_j}_2^2 + \alpha_j^2 \Big) \\
        & \hspace{15mm} = \begin{pmatrix}
            \displaystyle \min_{(v_i, w_i)\iPhat} \ell \bigg( \sum\iPhat D_i (X+\DeltaStar) (v_i - w_i), y \bigg) + \beta \sum\iPhat \big( \norm{v_i}_2 + \norm{w_i}_2 \big) \\[4mm]
            \ST \; (2 D_i - I_n) (X+\DeltaStar) v_i \geq 0, \; (2 D_i - I_n) (X+\DeltaStar) w_i \geq 0, \; \alliinPhat
        \end{pmatrix}.
    \end{aligned}
\end{equation*}

To address the feasibility issue, we can apply robust optimization techniques (\citep{boyd2004convex} section 4.4.2) and replace the constraints in (\ref{minimax2}) with robust convex constraints, which will lead to (\ref{eq:rob_gen_cvx}). 
Let $\big( (v_{\text{rob}_i}^\star, w_{\text{rob}_i}^\star)\iPhat, \Delta_{\text{rob}}^\star \big)$ denote an optimal point of (\ref{eq:rob_gen_cvx}) and let $(u_{\text{rob}_j}^\star, \alpha_{\text{rob}_j}^\star)\jmhs$ be the neural network weights recovered from $(v_{\text{rob}_i}^\star, w_{\text{rob}_i}^\star)\iPhat$ with (\ref{recover_weights}), where $\mhs$ is the number of nonzero weights.
In light of Lemma \ref{LEMMA:RECOVER}, since the constraints $(2 D_i - I_n) (X+\Delta) v_{\text{rob}_i}^\star \geq 0$ and $(2 D_i - I_n) (X+\Delta) w_{\text{rob}_i}^\star \geq 0$ for all $i\in[\widehat{P}]$ apply to all $X+\Delta\in\gU$, all $X+\Delta\in\gU$ satisfy the equality
\begin{align*}
    \displaystyle & \ell \bigg( \sum\iPhat D_i (X+\Delta) (v_{\text{rob}_i}^\star - w_{\text{rob}_i}^\star), y \bigg) + \beta \sum\iPhat \big( \norm{v_{\text{rob}_i}^\star}_2 + \norm{w_{\text{rob}_i}^\star}_2 \big) \\
    & \hspace{30mm} = \ell \bigg( \sum\jmhs \big( (X+\Delta) u_{\text{rob}_j}^\star \big)_+ \alpha_{\text{rob}_j}^\star, y \bigg) + \frac{\beta}{2} \sum\jmhs \big( \norm{u_{\text{rob}_j}^\star}_2^2 + \alpha_{\text{rob}_j}^{\star 2} \big).
\end{align*} 

Thus, since 
\begin{equation*}
    \displaystyle \Delta_{\text{rob}}^\star = \argmax\allDeltaU \ell \bigg( \sum\iPhat D_i (X+\Delta) (v_{\text{rob}_i}^\star - w_{\text{rob}_i}^\star), y \bigg) + \beta \sum\iPhat \big( \norm{v_{\text{rob}_i}^\star}_2 + \norm{w_{\text{rob}_i}^\star}_2 \big),
\end{equation*}
we have 
\begin{equation*}
    \displaystyle \Delta_{\text{rob}}^\star = \argmax\allDeltaU \ell \bigg( \sum\jmhs \big( (X+\Delta) u_{\text{rob}_j}^\star \big)_+ \alpha_{\text{rob}_j}^\star, y \bigg) + \frac{\beta}{2} \sum\jmhs \big( \norm{u_{\text{rob}_j}^\star}_2^2 + \alpha_{\text{rob}_j}^{\star 2} \big),
\end{equation*}
giving rise:
\begin{equation*}
    \begin{aligned}
        & \ell \bigg( \sum\iPhat D_i (X+\Delta_{\text{rob}}^\star) (v_{\text{rob}_i}^\star - w_{\text{rob}_i}^\star), y \bigg) + \beta \sum\iPhat \big( \norm{v_{\text{rob}_i}^\star}_2 + \norm{w_{\text{rob}_i}^\star}_2 \big) \\
        & \hspace{25mm} = \ell \bigg( \sum\jmhs \big( (X+\Delta_{\text{rob}}^\star) u_{\text{rob}_j}^\star \big)_+ \alpha_{\text{rob}_j}^\star, y \bigg) + \frac{\beta}{2} \sum\jmhs \big( \norm{u_{\text{rob}_j}^\star}_2^2 + \alpha_{\text{rob}_j}^{\star 2} \big) \\
        & \hspace{25mm} = \max\allDeltaU \ell \bigg( \sum\jmhs \big( (X+\Delta) u_{\text{rob}_j}^\star \big)_+ \alpha_{\text{rob}_j}^\star, y \bigg) + \frac\beta2 \sum\jmhs \big( \norm{u_{\text{rob}_j}^\star}_2^2 + \alpha_{\text{rob}_j}^{\star 2} \big) \\
        & \hspace{25mm} \geq \min_{(u_j, \alpha_j)\jmhs} 
        \begin{pmatrix}
            \displaystyle \max\allDeltaU \ell \bigg( \sum\jmhs \big( (X+\Delta) u_j \big)_+ \alpha_j, y \bigg) + \frac\beta2 \sum\jmhs \big( \norm{u_j}_2^2 + \alpha_j^2 \big)
        \end{pmatrix}
    \end{aligned}
\end{equation*}

Therefore, (\ref{eq:rob_gen_cvx}) is an upper bound to (\ref{robust_general}).

While there are an infinite number of points in the uncertainty set, which makes enumerating all points intractable, one can only enumerate all possible locations of the worst-case adversarial perturbation $\DeltaSStar$. For some loss functions, the possible locations are finitely many points. Moreover, note that the number of $D$ matrices is trivially upper-bounded by $2^n$.

\subsection{Proof of Corollary \ref{CORO:ROB_CONSTRAINT}} \label{sec:ROB_CONSTRAINT}
Define $E_i = 2 D_i - I_n$ for all $i\in[\widehat{P}]$.
Note that each $E_i$ is a diagonal matrix, and its diagonal elements are either -1 or 1. Therefore, for each $i\in[\widehat{P}]$, we can analyze the robust constraint $\min\allDeltaU E_i (X+\Delta) v_i \geq 0$ element-wise (for each data point).
Let $e_{ik}$ denote the $k^\text{th}$ diagonal element of $E_i$ and $\delta_{ik}^\top$ denote the $k^\text{th}$ element of $\Delta$ that appears in the $i^\text{th}$ constraint. We then have:
\begin{equation} \label{hinge_constraint_min}
    \begin{pmatrix}
        \displaystyle \min_{\norm{\delta_{ik}}_\infty \leq \epsilon} e_{ik} (x_k^\top + \delta_{ik}^\top) v_i
    \end{pmatrix} = \begin{pmatrix}
        \displaystyle e_{ik} x_k^\top v_i + \min_{\norm{\delta_{ik}}_\infty \leq \epsilon} e_{ik} \delta_{ik}^\top v_i
    \end{pmatrix} \geq 0
\end{equation}

The minima of the above optimization problems are achieved at $\delta_{ik}^{\star\star} = \epsilon \cdot \sgn(e_{ik} v_i) = \epsilon \cdot e_{ik} \cdot \sgn(v_i)$.

Note that as $\epsilon$ approaches 0, $\delta_{ik}^{\star\star}$ and $\Delta_{\text{rob}}^\star$ in Theorem \ref{THM:CVX_MINIMAX} both approach 0, which means that the gap between the convex robust problem (\ref{HINGE_ADV_D}) and the non-convex adversarial training problem (\ref{hinge_adv}) diminishes.
Plugging $\delta_k^{\star\star}$ into (\ref{hinge_constraint_min}) yields that
\begin{equation*}
\begin{aligned}
    \Big( e_{ik} x_k^\top v_i - \epsilon \norm{e_{ik} v_i}_1 \Big)
    = \Big( e_{ik} x_k^\top v_i - \epsilon \norm{v_i}_1 \Big) \geq 0.
\end{aligned}
\end{equation*}
Vertically concatenating $e_{ik} x_k^\top v_i - \epsilon \norm{v_i}_1 \geq 0$ for all $i\in[\widehat{P}]$ gives the vectorized representation $E_i X v_i - \epsilon \norm{v_i}_1 \geq 0$, which leads to (\ref{robust_constraint}).

Since the constraints on $w$ are exactly the same, we also have that $\min\allDeltaU E_i (X+\Delta) w_i \geq 0$ is equivalent to $E_i X w_i - \epsilon \norm{w_i}_1 \geq 0$ for every $i\in[\widehat{P}]$.

\subsection{Proof of Theorem \ref{LEMMA:INNER_MAX}} \label{sec:INNER_MAX}
The regularization term is independent from $\Delta$. Thus, it can be ignored for the purpose of analyzing the inner maximization. Note that each $D_i$ is diagonal, and its diagonal elements are either 0 or 1. Therefore, the inner maximization of (5) can be analyzed element-wise (cost of each data point).

The maximization problem of the loss at each data point is:
\begin{equation}\label{hinge_inner_max}
    \max_{\norm{\delta_k}_\infty \leq \epsilon} \bigg( 1 - y_k \sum\iP d_{ik} (x_k^\top+\delta_k^\top) (v_i - w_i) \bigg)_+ \\
\end{equation}
where $d_{ik}$ is the $k^\text{th}$ diagonal element of $D_i$ and $\delta_k^\top$ is the $k^\text{th}$ row of $\Delta$. One can write:
\begin{equation*}
\begin{aligned}
    & \max_{\norm{\delta_k}_\infty \leq \epsilon} \bigg( 1 - y_k \sum\iP d_{ik} (x_k^\top+\delta_k^\top) (v_i - w_i) \bigg)_+ \\
    & \hspace{35mm} = \bigg( \max_{\norm{\delta_k}_\infty \leq \epsilon} 1 - y_k \sum\iP d_{ik} (x_k^\top+\delta_k^\top) (v_i - w_i) \bigg)_+ \\
    & \hspace{35mm} = \bigg( 1 - y_k \sum\iP d_{ik} x_k^\top (v_i - w_i)
    - \min_{\norm{\delta_k}_\infty \leq \epsilon} \delta_k^\top y_k \sum\iP d_{ik} (v_i - w_i) \bigg)_+ .\\
\end{aligned}
\end{equation*}

The optimal solution to $\displaystyle \min_{\norm{\delta_k}_\infty \leq \epsilon} \delta_k^\top y_k \sum\iP d_{ik} (v_i - w_i)$ is $\displaystyle \delta_{\text{hinge}_k}^\star = - \epsilon \cdot \sgn \Big( y_k \sum\iP d_{ik} (v_i - w_i)^\top \Big)$, or equivalently: 
\begin{equation*} 
    \Delta_\text{hinge}^\star = - \epsilon \cdot \sgn \Big( \sum\iP D_i y (v_i - w_i)^\top \Big).
\end{equation*}

By substituting $\delta_{\text{hinge}_k}^\star$ into (\ref{hinge_inner_max}), the optimization problem (\ref{hinge_inner_max}) reduces to:
\begin{equation*}
\begin{aligned}
    & \bigg ( 1 - y_k \sum\iP d_{ik} x_k^\top (v_i - w_i)
    + \epsilon \bigg|\bigg| y_k \sum\iP d_{ik} (v_i - w_i)\bigg|\bigg| _1 \bigg)_+ \\
    & \hspace{35mm} = \bigg( 1 - y_k \sum\iP d_{ik} x_k^\top (v_i - w_i)
    + \epsilon |y_k| \bigg|\bigg| \sum\iP d_{ik} (v_i - w_i) \bigg|\bigg|_1 \bigg)_+ .
\end{aligned}
\end{equation*}

Therefore, the overall loss function is:
\begin{equation*}
    \frac{1}{n} \: \sum_{k=1}^n \bigg( 1 - y_k \sum\iP d_{ik} x_k^\top (v_i - w_i) + \epsilon |y_k| \bigg|\bigg| \sum\iP d_{ik} (v_i - w_i) \bigg|\bigg|_1 \bigg)_+ .
\end{equation*}

In the case of binary classification, $y = \{-1, 1\}^n$, and thus $|y_k| = 1$ for all $k\in[n]$. Therefore, the above is equivalent to
\begin{equation} \label{hinge_obj}
    \frac{1}{n} \: \sum_{k=1}^n \bigg( 1 - y_k \sum\iP d_{ik} x_k^\top (v_i - w_i) + \epsilon \bigg|\bigg| \sum\iP d_{ik} (v_i - w_i) \bigg|\bigg|_1 \bigg)_+
\end{equation}
which is the objective of (\ref{HINGE_ADV_D}). This completes the proof.

\subsection{Proof of Theorem \ref{THEOREM:ROBUST_SOCP}}\label{sec:ROBUST_SOCP}
We first exploit the structure of (\ref{sl_minimax_convex}) and reformulate it as the following robust second-order cone program (SOCP):
\vspace{-2mm}
\begin{align}\label{SOCP2}
         &\min_{(v_i, w_i, b_i, c_i)\iPhat, a} a + \beta\sum\iPhat (b_i+c_i) \\
        &\ST \;  (2 D_i - I_n) X v_i \geq \epsilon \norm{v_i}_1, \;\;
        (2 D_i - I_n) X w_i \geq \epsilon \norm{w_i}_1, \;\;
        \norm{v_i}_2\leq b_i, \;\; \norm{w_i}_2\leq c_i, \;\; \alliinPhat \nonumber \\
        &\qquad \max\allDelta \begin{Vmatrix} \begin{bmatrix} \sum\iPhat 
        D_i(X+\Delta)(v_i-w_i) - y \\ 2a-\frac{1}{4} \end{bmatrix} \end{Vmatrix}_2 \leq 2a+\tfrac{1}{4}, \qquad \alliinPhat. \nonumber
\end{align}

Then, we need to establish the equivalence between (\ref{SOCP2}) and (\ref{SOCP3}). To this end, we consider the constraints of (\ref{SOCP2}) and argue that these can be recast as the constraints given in (\ref{SOCP3}).
One can write:
\begingroup
\allowdisplaybreaks
\begin{align*}
    & \max\allDelta \Bigg|\Bigg| \begin{bmatrix} \sum\iPhat D_i(X+\Delta)(v_i-w_i) - y \\ 2a-\frac{1}{4} \end{bmatrix} \Bigg|\Bigg|_2 \leq 2a+\frac{1}{4} \\
    \Longleftrightarrow &
    \max_{\norm{\delta_k}_\infty\leq \epsilon,\ \forall k\in[n]}  \begin{Vmatrix} \begin{bmatrix} 
        \sum\iPhat d_{i1}(x_1^\top-\delta_1^\top)(v_i-w_i) - y_1 \\ \sum\iPhat d_{i2}(x_2^\top-\delta_2^\top)(v_i-w_i) - y_2\\
        \vdots \\
        \sum\iPhat d_{in}(x_n^\top-\delta_n^\top)(v_i-w_i) - y_n\\
        2a-\frac{1}{4} 
    \end{bmatrix} \end{Vmatrix} _2 \leq 2a+\frac{1}{4} \\
    \Longleftrightarrow & 
    \max_{\norm{\delta_k}_\infty \leq \epsilon,\ \forall k\in[n]} \bigg(\sum_{k=1}^n \Big(\sum\iPhat d_{ik}(x_k^\top-\delta_k^\top)(v_i-w_i) - y_k\Big)^2 + \Big(2a-\frac{1}{4}\Big)^2\bigg)^\frac{1}{2} \leq 2a+\frac{1}{4}
\end{align*}
\endgroup
where $d_{ik}$ is the $k^\text{th}$ diagonal element of $D_i$ and $\delta_k^\top$ is the $k^\text{th}$ row of $\Delta$. The above constraints can be rewritten by introducing slack variables $z\in\mathbb{R}^{n+1}$ as
\begin{equation*}
\begin{aligned}
  & \textstyle z_k \geq \Big| \sum\iPhat d_{ik} x_k^\top (v_i-w_i) - y_k \Big| + \epsilon \Big|\Big| \sum\iPhat d_{ik} (v_i-w_i) \Big|\Big|_1, \ \forall k \in [n] \\
  & z_{n+1} \geq \big| 2a-\tfrac{1}{4} \big|, \quad \norm{z}_2 \leq 2a+\tfrac{1}{4}.
\end{aligned}
\end{equation*}
\hfill $\blacksquare$

\subsection{Proof of Lemma \ref{LEMMA:RED_D}} \label{sec:RED_D}
According to \citep{pmlr-v119-pilanci20a}, recovering the neural network weights by plugging (\ref{recover_weights}) in (\ref{theorem_5}) leads to
\begin{equation*}
\begin{aligned}
    q^\star = & \min_{(v_i, w_i)\iP} \ell \Bigg( \sum\iP D_i X (v_i - w_i), y \Bigg) + \beta \sum\iP \Big( \norm{v_i}_2 + \norm{w_i}_2 \Big) \\
    = & \min_{(u_j, \alpha_j)\jms} \ell \Bigg( \sum\jms (X u_j)_+ \alpha_j, y \Bigg) + \frac{\beta}{2} \sum\jms \Big( \norm{u_j}_2^2 + \alpha_j^2 \Big)
\end{aligned}
\end{equation*}

\newcommand{\vits}{\widetilde{v}_i^\star}
\newcommand{\wits}{\widetilde{w}_i^\star}
\newcommand{\util}{\widetilde{u}}
\newcommand{\altil}{\widetilde{\alpha}}
\newcommand{\mts}{\widetilde{m}^\star}
Similarly, we can recover the neural network weights from the solution $(\widetilde{v}_i^\star, \widetilde{w}_i^\star)\itildeP$ of (\ref{redundant_D}) using:
\begin{equation}\label{recover_weights_1} 
    \begin{aligned}
        (\util_{j_{1 i}}, \altil_{j_{1 i}}) = \rbr*{\dfrac{\vits}{\sqrt{\norm{\vits}_2}}, \sqrt{\norm{\vits}_2}}, \quad 
        (\util_{j_{2 i}}, \altil_{j_{2 i}}) = \rbr*{\dfrac{\wits}{\sqrt{\norm{\wits}_2}}, -\sqrt{\norm{\wits}_2}}, \quad \forall i \in [\widetilde{P}].
    \end{aligned}
\end{equation}

Unlike (\ref{recover_weights}), zero weights are not discarded in (\ref{recover_weights_1}). For simplicity, we use $\util_1, \dots, \util_{\mts}$ to refer to the hidden layer weights and use $\altil_1, \dots, \altil_{\mts}$ to refer to the output layer weights recovered using (\ref{recover_weights_1}). Since $(\vits, \wits)\itildeP$ is a solution to (\ref{redundant_D}), it satisfies $(2 D_i - I_n) X \vits \geq 0$ and $(2 D_i - I_n) X \wits \geq 0$ for all $i \in [\widetilde{P}]$. Thus, we can apply Lemma \ref{LEMMA:RECOVER} to obtain:
\begin{align*}
    \widetilde{q}^\star = & \ell \bigg(\sum\itildeP D_i X (\vits - \wits), y \bigg) + \beta \sum\itildeP \Big(\norm{\vits}_2 + \norm{\wits}_2 \Big) \\
    = & \ell \bigg(\sum_{j=1}^{\mts} (X \util_j^\star)_+ \alpha_j, y \bigg) + \frac{\beta}{2} \sum_{j=1}^{\mts} \Big( \norm{\util_j^\star}_2^2 + \altil_j^{\star 2} \Big) \\
    \geq & \min_{(u_j, \alpha_j)_{j=1}^{\mts}} \ell \bigg(\sum_{j=1}^{\mts} (X u_j)_+ \alpha_j, y \bigg) + \frac{\beta}{2} \sum_{j=1}^{\mts} \Big( \norm{u_j}_2^2 + \alpha_j^2 \Big)
\end{align*}
Since $\widetilde{P} \geq P$, $m^\star \leq 2 P$ and $\mts = 2 \widetilde{P}$, we have $\widetilde{m}^\star \geq m^\star$. Therefore, according to Section 2 and Theorem 6 of \citep{pmlr-v119-pilanci20a}, we have
\begin{equation*}
\begin{aligned}
    q^\star = & \min_{(u_j, \alpha_j)\jms} \ell \bigg( \sum_{j=1}^{m^\star} (X u_j)_+ \alpha_j, y \bigg) + \frac{\beta}{2} \sum\jms \Big( \norm{u_j}_2^2 + \alpha_j^2 \Big) \\
    = & \min_{(u_j, \alpha_j)_{j=1}^{\widetilde{m}^\star}} \ell \bigg( \sum_{j=1}^{\widetilde{m}^\star} (X u_j)_+ \alpha_j, y \bigg) + \frac{\beta}{2} \sum_{j=1}^{\widetilde{m}^\star} \Big( \norm{u_j}_2^2 + \alpha_j^2 \Big) \\
    \leq & \ \widetilde{q}^\star
\end{aligned}
\end{equation*}

The above inequality shows that a neural network with more than $m$ neurons in the hidden layer will yield the same loss as the neural network with $m$ neurons when optimized.

Note that (\ref{redundant_D}) can always attain $q^\star$ by simply plugging in the optimal solution of (\ref{theorem_5}) and assigning 0 to all other additional $v_i$ and $w_i$, implying that $q^\star \geq \widetilde{q}^\star$. Since $q^\star$ is both an upper bound and a lower bound on $\widetilde{q}^\star$, we have $\widetilde{q}^\star = q^\star$, proving that as long as all matrices in $\mathcal{D}$ are included, the existence of redundant matrices does not change the optimal objective value.

\subsection{$\ell_p$ norm-bounded perturbation set for hinge loss} \label{lpnorm}
Theorem \ref{LEMMA:INNER_MAX} can be extended to the following $\ell_p$ norm-bounded perturbation set: 
\begin{equation*}
    \widetilde{\gX} = \big\{ X + \Delta \in \sR^{n \times d} \ \big| \ \Delta = [\delta_1^\top; \ldots ; \delta_n^\top], \ \norm{\delta_k}_p \leq \epsilon, \ \forall k \in [n] \big\}
\end{equation*}
In the case of performing binary classification with a hinge-lossed neural network, the convex adversarial training problem then becomes:
\begin{align}\label{HINGE_ADV_D_lp}
    \min_{(v_i, w_i)\iPhat} & 
    \begin{pmatrix}
        \displaystyle \frac{1}{n} \: \sum_{k=1}^n \bigg( 1 - y_k \sum\iP d_{ik} x_k^\top (v_i - w_i) + \epsilon \bigg| \bigg| \sum\iP d_{ik} (v_i - w_i) \bigg| \bigg|_{p*} \bigg)_+ \\[1mm]
        \hfill + \beta \sum\iPhat \Big( \norm{v_i}_2 + \norm{w_i}_2 \Big)
    \end{pmatrix} \\
    \ST & \quad (2 D_i - I_n) X v_i \geq \epsilon \norm{v_i}_{p*}, \quad (2 D_i - I_n) X w_i \geq \epsilon \norm{w_i}_{p*}, \quad \alliinPhat \nonumber
\end{align}
where $D_1, \dots, D_{\hat{P}}$ are all distinct diagonal matrices associated with $\diag([X u \geq 0])$ for all possible $u \in \mathbb{R}^d$ and all $X+\Delta$ at the \textit{boundary} of $\widetilde{\gX}$. Moreover, $\norm{\cdot}_{p*}$ is the dual norm of $\norm{\cdot}_p$.

\end{document}